\documentclass[preprint,12pt]{elsarticle}

\usepackage{amssymb}
\usepackage{amsmath}

\usepackage{url}
\usepackage{amsmath}
\usepackage{graphicx}
\usepackage{epsfig,amsmath,amsfonts}
\usepackage{multirow}
\usepackage{float}
\usepackage{xcolor}
\usepackage{caption}
\usepackage{subfigure}
\usepackage{placeins}

\newcommand{\eq}[1]{Eq.\ (\ref{#1})}

\newcommand{\ZL}[1]{\textcolor[rgb]{0,0.0,0}{#1}}

\begin{document}

\begin{frontmatter}



\title{Hybrid Adaptive Modeling using Neural Networks Trained with Nonlinear Dynamics Based Features}


\author{Zihan Liu} 
\author{Prashant N. Kambali}
\author{C. Nataraj}

\affiliation[]{organization={Villanova Center for Analytics of Dynamic Systems},\\
            addressline={Villanova University},
            city={Villanova},
            postcode={19085},
            state={PA},
            country={USA}\\
            http://vcads.org\\
        (zliu5; prashant.kambali; nataraj)@villanova.edu}

\begin{abstract}
\ZL{Accurate models are essential for design, performance prediction, control, and diagnostics in complex engineering systems. Physics-based models excel during the design phase but often become outdated during system deployment due to changing operational conditions, unknown interactions, excitations, and parametric drift. While data-based models can capture the current state of complex systems, they face significant challenges, including excessive data dependence, limited generalizability to changing conditions, and inability to predict parametric dependence.} This has led to combining physics and data in modeling, termed physics-infused machine learning, often using numerical simulations from physics-based models. This paper introduces a novel approach that departs from standard techniques by uncovering information from nonlinear dynamical modeling and embedding it in data-based models. The goal is to create a hybrid adaptive modeling framework that integrates data-based modeling with newly measured data and analytical nonlinear dynamical models for enhanced accuracy, parametric dependence, and improved generalizability. By explicitly incorporating nonlinear dynamic phenomena through perturbation methods, the predictive capabilities are more realistic and insightful compared to knowledge obtained from brute-force numerical simulations. In particular, perturbation methods are utilized to derive asymptotic solutions which are parameterized to generate frequency responses. Frequency responses provide comprehensive insights into dynamics and nonlinearity which are quantified and extracted as high-quality features. A machine-learning model, trained by these features, tracks parameter variations and updates the mismatched model. \ZL{The results demonstrate that this adaptive modeling method outperforms numerical gray box models in prediction accuracy and computational efficiency.} 

\end{abstract}

\begin{graphicalabstract}
\includegraphics[scale=0.4]{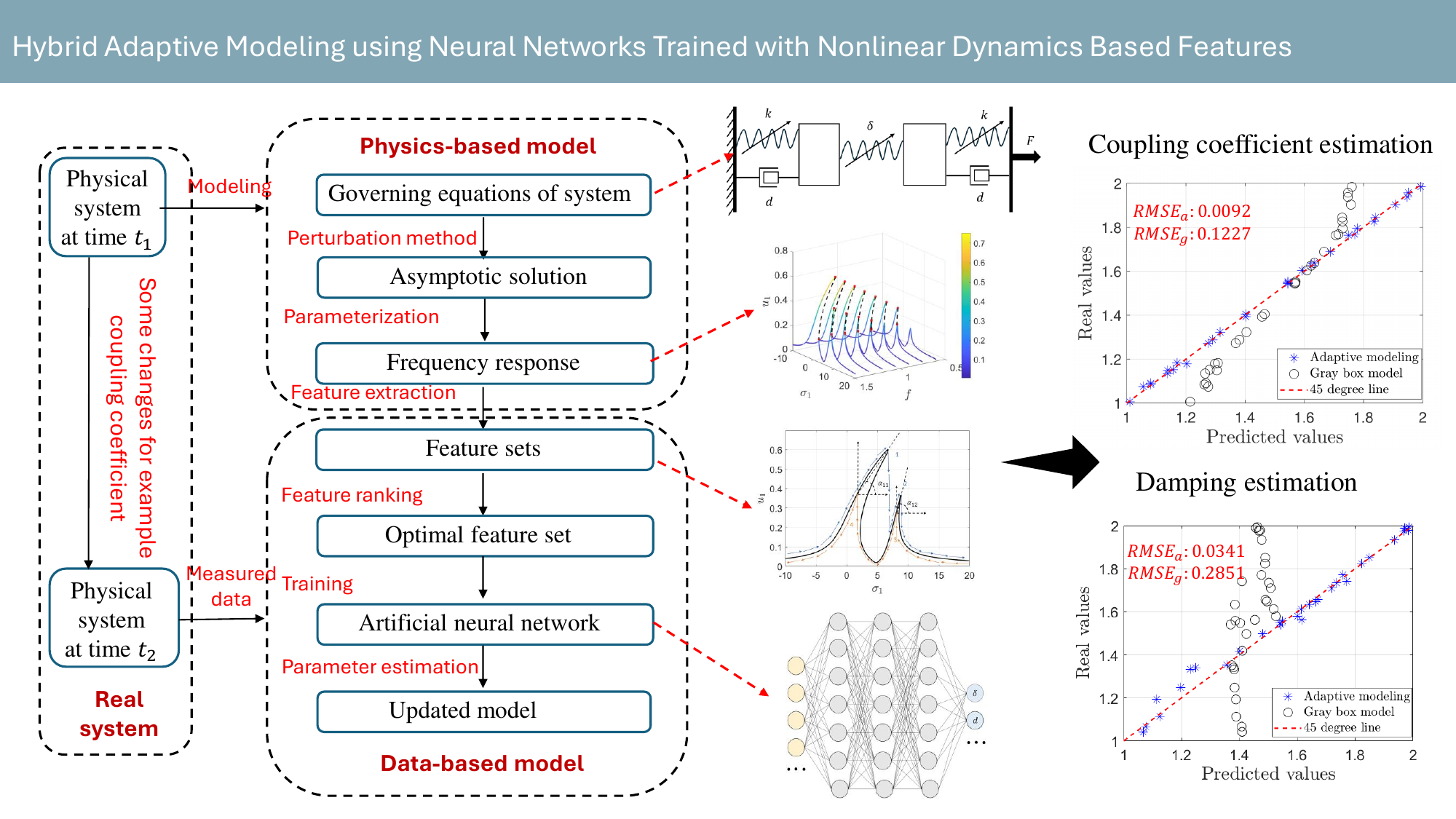}
\end{graphicalabstract}

\begin{highlights}
\item The proposed method combines physics-based modeling and data-based modeling to model complex nonlinear systems and adapt to parameter changes.

\item The physical model is parameterized to synthesize data for model training in the absence of real data.

\item Features extracted from nonlinear phenomena, captured by analytical solutions, are highly effective for tracking parameter variations that have an impact on related nonlinear phenomena.

\item Feature ranking and selection techniques are employed to form an appropriate feature set for model training.

\item The nonlinear dynamics-based features enhance the generalization and robustness of the proposed model compared to widely used gray box modeling in parameter identification.
\end{highlights}


\end{frontmatter}

\section{Introduction}
\label{sec:Introduction}
The evolution of systems from localized, regulated entities into interconnected, intricate large networks has inevitably introduced challenges to accurate modeling. As systems grow in complexity, the task of modeling becomes increasingly error-prone. \ZL{To address this, the integration of self-adaptation capabilities, adjusting their behavior to operational condition changes, uncertainty, nonlinearity, and various application environments to maintain their functionality and performance~\cite{krupitzer2024rethinking,yang2023novel}, has emerged as a viable solution.} An adaptive system can automatically modify itself in response to changes in its operating environment and model drift. \ZL{For achieving adaptation, one of the fundamental system properties is self-awareness~\cite{salehie2009self}, which refers to the ability to monitor resources, state and behavior. Hence, the newly measured data is critical to indicate state changes in the system. More importantly, it is essential to clearly identify the type and impact of these changes to determine whether adaptation is necessary. These decisions are grounded in prior knowledge of the system: models, rules, policies, goals and utility~\cite{lalanda2013autonomic}. The following question is `What kind of change is needed?' Parameter adaptation is a straightforward approach.} In a well-developed physics-based model, unknown parameters or real-time varied parameters (e.g., excess loads, external disturbance, changes in ambient conditions) are identified and updated by incorporating the measured dynamic data. \ZL{However, for complex realistic systems with nonlinear changes and effects that exceed the established model's scope, additional components or environmental modifications may be required to prevent performance degradation. In these cases, data-based modeling is frequently employed in adaptive systems due to its flexibility and ability to learn nonlinear patterns.} 

\ZL{Krupitzer~{et al.}~\cite{krupitzer2015survey,krupitzer2020overview} study engineering approaches of self-adaptive systems with notable references to specific applications in control~\cite{fan2018adaptive,Cartes2005}, biomedical~\cite{del2011adaptive}, mechanical~\cite{de2003towards,frey2012architectural}, and transportation~\cite{tomforde2014incremental,fredericks2019planning}.} In such adaptive systems, the accuracy and robustness of the adaptive model play a crucial role. The decline in system performance can be attributed to disparities within the model. Factors like wear, fatigue, and varying operating circumstances such as coupling with other mechanical components can lead to model drift and mismatch. As a result, developing an adaptive modeling technique for tracking system fluctuations is of great significance. It is necessary to note that system variations with time are usually orders of magnitude larger than the system dynamic response time.

Physics-based modeling (PBM) is a traditional modeling approach that has acquired reasonable verisimilitude with real behaviors by employing traditional mathematical models and computer codes~\cite{velten2024mathematical}. In general,  PBM is based on physics or well-established rules, which are capable of qualitatively capturing some observable effects and providing useful insights into underlying causalities. The limitations of PBM derive from the fact that many realistic physical systems are getting increasingly complicated, beyond what can be adequately modeled with the physics we know. Recently, as machine learning, sensor technology, computer hardware, and big data continue to progress, the system's uncertainties and unpredictability become more discernible in the acquired data. These have inspired a novel approach to model physical systems, namely, Data-based Modeling (DBM)~\cite{bishnu2023computational}. Typically, the DBM approach uses observed data to learn a pattern that is used to predict system response for future inputs. Because DBM is heavily dependent on the quality and quantity of collected data, as well as the type of the system, effective, accurate and sufficient data acquisition is critical for the applications of the DBM approach. An intrinsic shortcoming of DBM is that it cannot be easily parameterized, or extended to situations that the model has not been exposed to. System deterioration, environmental changes, and operation condition changes, all of which are typical in practice, can jeopardize the validity and generalizability of the DBM. Additionally, data-based models are limited in revealing the physical causality underlying observed dynamics. As a result, they are incompetent to predict the parametric dependence of various nonlinear phenomena.

Adaptive models rely on prior knowledge to construct a physical model meanwhile utilizing newly measured data from altered systems. This inherent characteristic suggests the combination of PBM and DBM. \ZL{Dong~{et al.}\cite{dong2024robust} proposed a nonlinear adaptive line-of-sight guidance algorithm in a light autonomous underwater vehicle system (LAUV) to update the varying attack angle for adaptation to perturbations.} Asad~{et al.}~\cite{asad2017hybrid} developed a Hybrid Genetic Algorithm by combining Genetic Algorithms and traditional optimization methods to reduce the mismatches between the model and the real system for an air-conditioning set-up. Ni~{et al.}~\cite{ni2016adaptive} proposed an adaptive state-space model for predicting the remaining useful life of a planetary gearbox. In order to model both slow dynamics and long-term dynamics of power systems, Stajcar~{et al.}~\cite{stajcar2016adaptive} introduced an adaptive modeling framework for dynamics of multiple time frames. Sands~\cite{sands2017nonlinear} proposed a methodology using a DC motor with a minimum-phase mathematical model controlled by a self-tuning regulator without a model pole cancellation. Elkhodary~{et al.}~\cite{elkhodary2010fusion} presented a Feature-Oriented Self-Adaptation framework to reduce the effects of uncertainty by learning the impacts of adaptation decisions on the system’s goals.
Hallsteinsen~{et al.}~\cite{hallsteinsen2012development} developed a comprehensive software framework that operates in ubiquitous and dynamic computing environments and adapts to changes in the environment. Amin~{et al.}~\cite{taghieh2022novel} proposed an interval type-3 fuzzy logic system to circumvent the shortcomings of fuzzy-based controllers in terms of approximating uncertainty and identifying nonlinear complex systems.

Recently, machine learning has picked up momentum as a relatively new technique in DBM. Therefore, there is a trend to apply machine learning frameworks to adaptive systems. Han~{et al.}~\cite{han2016adaptive} employed an adaptive second-order algorithm to train the Fuzzy Neural Network for fast and robust convergence. \ZL{Meng~{et al.}~\cite{meng2023novel} proposed an adaptive Kriging model to approximate the actual performance function of a reliability-based design optimization problem. The effectiveness is proven in the optimal design of an offshore wind turbine monopile. Yang~{et al.}~\cite{yang2024novel} developed a novel learning function to find optimal samples for improving the accuracy of the surrogate model based on the fundamental assumptions of spatial correlations and stationarity.} Lin~{et al.}~\cite{lin2017adaptive} identified a nonlinear system with parametric uncertainty using the Takagi-Sugeno fuzzy model and then suggested an adaptive parameter estimator to track unknown nonlinear system parameters. Jalali~{et al.}~\cite{Jalali2010} developed an Adaptive Neuro-Fuzzy Inference System structure to identify the heart rate baroreflex mechanism. Sitompul~\cite{sitompul2014neural} modified the traditional neural network by expanding parameters between the output layer and hidden layer and connecting the output and input to add external recurrence. This structure is integrated into an adaptive scheme to model systems with changing properties or various operating conditions. Liu~{et al.}~\cite{liu2019adaptive} used an adaptive neural network to approximate the unknown mass of the car body of an Active Suspension System with time-varying vertical displacement and speed constraints.

\begin{figure}[ht]
\centering
\includegraphics[scale=0.5]{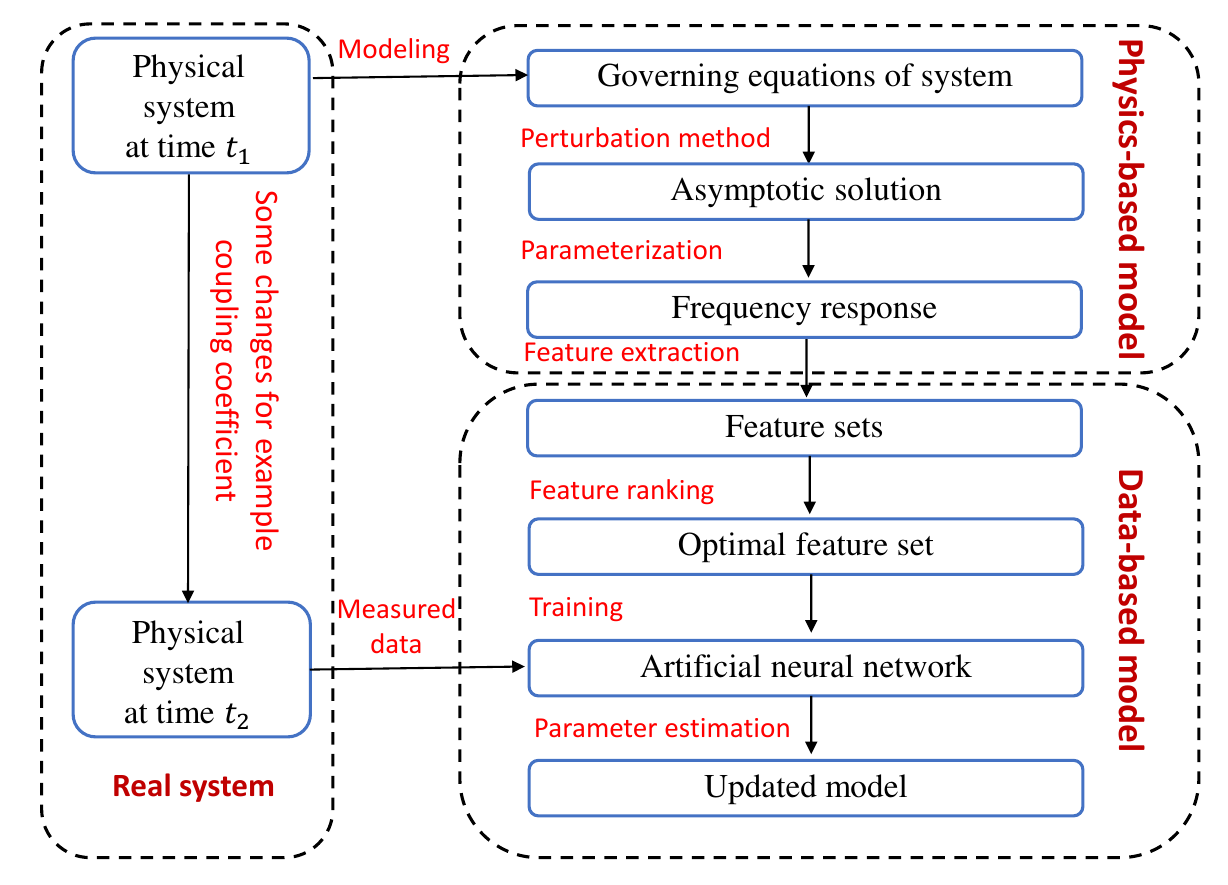}
  \caption{The overview of the adaptive modeling approach}
\label{fig:overview}
\end{figure}


\subsection*{Proposed Innovation}
\label{subsec:proposed innovation}
The above-described methods which use neural networks to derive adaptive systems are effective to varying degrees. However, there are still difficulties with generalizability which is a common challenge in machine learning algorithms. \ZL{This occurs when a model is trained on a specific dataset and performs well on that dataset but fails to generalize to new, unseen data~\cite{simon2009neural}.} So, we are interested in solving this generalizability problem. For this reason, a novel hybrid modeling approach is developed for tracking system parameter variations by parameterizing the system in a physics-based model. \ZL{As a result, the training dataset generated from the physics-based model encompasses a spectrum of possible scenarios,  effectively covering most real operational conditions.} It should be noted that standard numerical simulation of physics-based models is normally employed in physics-informed machine learning. Our lab has been exploring its application in a variety of contexts~\cite{liu2023diagnostic,haj2018rolling,abbasi2022adaptive,abbasi2022physics}. Here, we depart from these standard approaches and embark on a new experiment. It is well known in the nonlinear dynamics literature that analytical and semi-analytical methods reveal parametric dependence of various nonlinear phenomena and other intrinsically insightful information much more readily than brute force numerical simulations. Hence, the question we ask in the current paper is the following: how do we exploit this unique aspect of analytical methods to help arrive at effective adaptive modeling approaches? To explore answers to this question, we have picked a well-known analytical method, the method of multiple scales (MMS). In addition, we pick an example nonlinear system that is not so simple that it would not enable a thorough investigation nor so complex that it would mire us in numerical details and defy detailed analysis. A set of coupled Duffing oscillators serves this purpose quite well. In addition to its intrinsic value as a prototypical example, it is also a candidate for the description of various physical and electro-mechanical systems~\cite{jothimurugan2016multiple,woafo1998dynamics}.

Our overall approach is as follows and is illustrated in Fig.~\ref{fig:overview}.
\begin{itemize}
\item We obtain frequency responses for the above system by varying parameters that define excitation, nonlinearity, dissipation and coupling. As is well known, frequency responses are more informative than time responses when it comes to tracking parameter changes. Some important dynamical features of the system, such as jump phenomena and bifurcation, are more apparent in frequency responses. Furthermore, these characteristics are easier to extract when using the frequency response to construct the feature space.
\item The synthetic data (generated analytically as explained above) is generated for different scenarios by varying damping and coupling coefficients with random load to mimic the real-world experimental data. Additionally, we also added noise to represent the uncertainties in the real world.
\item Depending on the parameters we wish to predict, we employ Mutual Information (MI) to rank all features and form an appropriate feature set.
\item These features are then fed into an Artificial Neural Network (ANN) to build a data-based model for system identification and parameter updating.
\end{itemize}

To reiterate, our focus is to derive a model that adapts to changing conditions; in this case, we limit ourselves to when the system parameters change. [Future papers will focus on other kinds of changes.] Our objective in this paper is hence to identify and track selected parameters in the coupled oscillators 
under a variety of real-world scenarios.

The rest of this paper is organized as follows: Section \ref{se:Physics-based Model} introduces the perturbation method applied in this paper and provides the details. Section \ref{se:data_based modeling} presents the feature extraction process, feature ranking and the training and fine-tuning of an Artificial Neural Network (ANN). In Section \ref{se:simulation and comparison}, the proposed method is applied in more realistic situations and compared with a standard gray box model. Finally, Section \ref{se:conclusion} summarizes and concludes the paper.

\section{Physics-based Model}
\label{se:Physics-based Model}
\subsection{Asymptotic analysis}
We consider a system of two coupled Duffing oscillators with sinusoidal excitation applied to the first oscillator. The non-dimensional equations are given as follows~\cite{jothimurugan2016multiple,woafo1998dynamics,salas2022analytical}.

\begin{eqnarray}
\begin{aligned}
x_1''+\omega_{0}^2x_1+d x_1'+\beta{x_{1}^3}+\delta(x_1-x_2)  &=  F\cos\Omega t  \\
x_2''+\omega_{0}^2x_2+d x_2'+\beta{x_{2}^3}+\delta(x_2-x_1)  &=  0,
\label{eq:non-dimention}
\end{aligned}
\end{eqnarray}

where $x_{1}$ and $x_{2}$ represent the displacement of the first and second oscillators respectively. $'$ denotes the derivative with respect to time, $\omega_{0}$ denotes the undamped uncoupled natural frequency, $d$ is the damping coefficient, $\beta$ is the coefficient of the nonlinear term, $\delta$ is the coupling coefficient, $F$ is the amplitude of excitation, and $\Omega$ is the frequency of excitation. Note that each of the oscillators has the standard cubic nonlinearity in its stiffness term but the coupling and the damping terms are linear.

The method of multiple scales (MMS)~\cite{kambali2017nonlinear,nayfeh2011introduction} is applied to solve \eq{eq:non-dimention} assuming the solutions as functions of multiple time scales, $T_0=t$, $T_1 = \epsilon t$.
\begin{eqnarray}
\begin{aligned}
x_1&=x_{10}(T_0,T_1) + \epsilon x_{11}(T_0,T_1) \\
x_2 &= x_{20}(T_0,T_1) + \epsilon x_{21}(T_0,T_1),
\label{eq:A_displacement}
\end{aligned}
\end{eqnarray}
where $\epsilon$ is a small positive dimensionless number. The derivative terms with respect to $t$ can therefore be defined as follows in terms of new time scales.

\begin{eqnarray}
\begin{aligned}
			\frac{d}{dt}=& D_{0} + \epsilon D_{1} + {\epsilon}^2 D_{2} +...\\
			\frac{d^2}{dt^2}=&  (D_{0} + \epsilon D_{1} + {\epsilon}^2 D_{2} +...)^2\\
			=& D_{0}^2 + 2 \epsilon D_{0} D_{1} + \epsilon^2(2D_{0}D_{2} + D_{1}^2) + ...,
			\label{eq:chain rule}
   \end{aligned}
\end{eqnarray}
	where
	\begin{equation}
		D_{n} = \frac{\partial}{\partial T_{n}}
		\label{eq;chain rule notation}
	\end{equation}

The damping, excitation, and nonlinear terms are then rescaled as $d=\epsilon d$, $F=\epsilon f$, $\beta = \epsilon \beta$. Substituting \eq{eq:A_displacement}, and \eq{eq:chain rule} into \eq{eq:non-dimention} and comparing different powers of $\epsilon$ up to the second order yields the following equations.

\begin{eqnarray}
\hspace*{-1.5cm}
 O(\epsilon^0):&&\nonumber\\
&&D_0^{2} x_{10} + \omega_0^{2}x_{10} + \delta x_{10} - \delta x_{20}=0\nonumber\\
&&D_0^{2} x_{20} + \omega_0^{2}x_{20} + \delta x_{20} - \delta x_{10}=0
\label{eq:order0}
\end{eqnarray}

\begin{eqnarray}
O(\epsilon^1):&&\nonumber\\
&&D_0^{2} x_{11} + \omega_0^{2}x_{11} + \delta x_{11} - \delta x_{21} =\nonumber\\&&-2D_0 D_1 x_{10} - d D_0 x_{10} - \beta x_{10}^{3} + f \cos \Omega t\nonumber\\
&&D_0^{2} x_{21} + \omega_0^{2}x_{21} + \delta x_{21} - \delta x_{11} =\nonumber\\&&-2D_0 D_1 x_{20} - d D_0 x_{20} - \beta x_{20}^{3}
\label{eq:order1}
\end{eqnarray}

The final solution is obtained by the usual method of multiple scales and is given as follows, the detailed derivation of which is given in Appendix (\eq{eq:Matrix of order0} -- \eq{eq:angle sum}).

\begin{eqnarray}
\begin{aligned}
x_{10} &=  c_1\cos{(\omega_1 t + \sigma_1 T_1)}  + c_2\sin{(\omega_1 t + \sigma_1 T_1)} \\
x_{20} &=  c_3\cos{(\omega_1 t + \sigma_1 T_1)} + c_4\sin{(\omega_1 t + \sigma_1 T_1)},
\label{eq:constant simplify}
\end{aligned}
\end{eqnarray}
where $c_1 = a_1\cos{\gamma_1} + a_2\cos{\gamma_2}$, $c_2 = a_1\sin{\gamma_1} + a_2\sin{\gamma_2}$, $c_3 = a_1\cos{\gamma_1} - a_2\cos{\gamma_2}$, $c_4 = a_1\sin{\gamma_1} - a_2\sin{\gamma_2}$. \ZL{$a_1$ and $a_2$ are introduced in \eq{eq:solution1_order0} for assuming the analytical solutions of $x_{10}$ and $x_{20}$, $\gamma_1$ and $\gamma_2$ are introduced in \eq{eq:modulation2} for conveniently expressing their phases.} \eq{eq:constant simplify} is further simplified and the results are given as follows.

\begin{eqnarray}
\begin{aligned}
x_{10} &= u_1 \cos{((\omega_1 + \epsilon \sigma_1) t - \Gamma_1)}\\
x_{20} &= u_2 \cos{((\omega_1 + \epsilon \sigma_1) t - \Gamma_2)},
\end{aligned}
\label{eq:solution1_order0_combination}
\end{eqnarray}

\noindent where $u_1 = \sqrt{c_1^2 + c_2^2}$, $u_2 = \sqrt{c_3^2 + c_4^2}$, are the amplitudes of $x_{10}$ and $x_{20}$ respectively; $\Gamma_1 = \arctan(\frac{c_2}{c_1})$,  $\Gamma_2 = \arctan(\frac{c_4}{c_3})$, are the phases of $x_{10}$ and $x_{20}$.

Choosing $\omega_0 = 1.0$, $\epsilon = 0.1$, $\sigma_2$ is calculated as $7.3$ according to \eq{eq:detuning}. \ZL{The analytical continuation theory~\cite{gamelin2003complex} is leveraged to parameterize the system’s behavior across a range of parameters. We employed the tool package MATCONT~\cite{dhooge2003matcont}, using the analytical continuation theory to trace the solutions of nonlinear dynamical systems along parameters variation, to generate the various frequency responses of the coupled oscillators which are illustrated in Figs.~\ref{fig:3d_delta}--\ref{fig:3d_f}
with various values of $\delta$, $\beta$, $d$, and $f$.} Note that dynamic characteristics like the jump phenomenon, bifurcation points, and bi-stability demonstrated in the frequency response curves are used as features for parameter estimation in the sequel.

\subsection{Parameterization}
\label{se:parameterization}
\begin{figure}[!ht]
\centering
  \subfigure[]{\includegraphics[scale=.4]{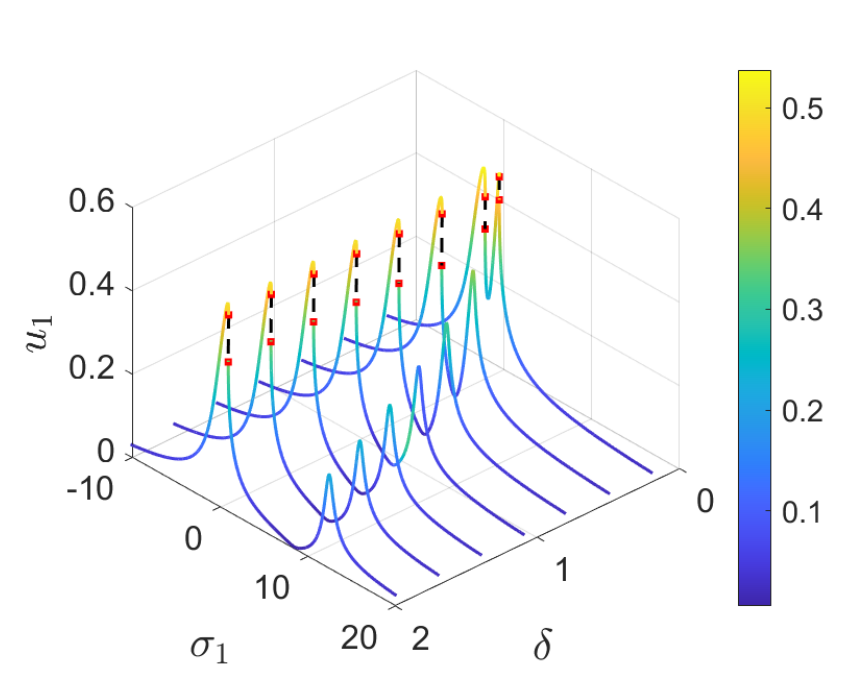}}
  \subfigure[]{\includegraphics[scale=.4]{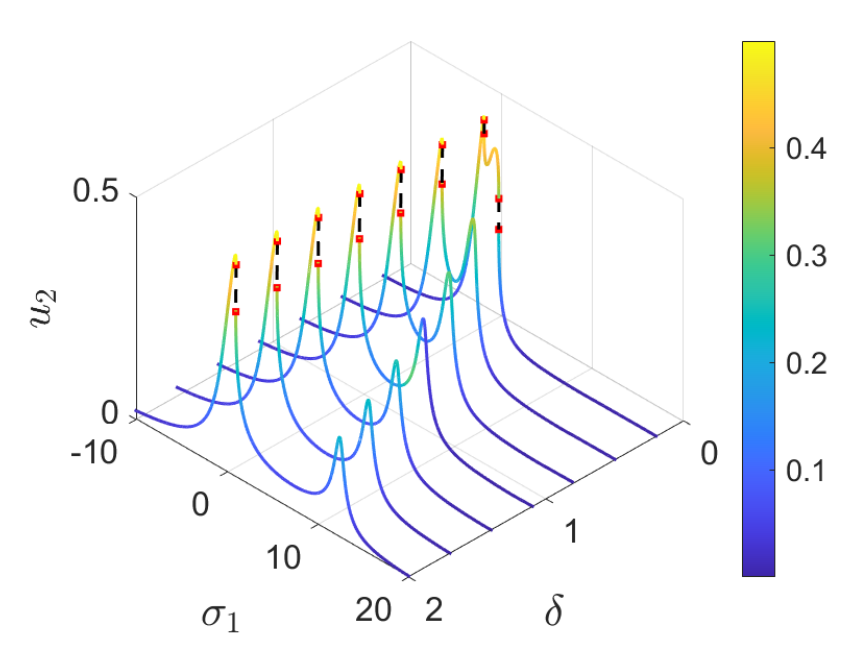}}
\caption{Frequency response of (a) oscillator $X$ and (b) oscillator $Y$ versus coupling coefficient $\delta$ and detuning parameter $\sigma_{1}$. For this case, $f = 1.0$, $d = 1.0$, $\epsilon = 0.1$, $\sigma_2 = 7.3$, $\beta = 10$ and $\delta$ ranging from $0.2$ to $2.0$.}
\label{fig:3d_delta}
\end{figure}

\begin{figure}[!ht]
\centering
  \subfigure[]{\includegraphics[scale=.4]{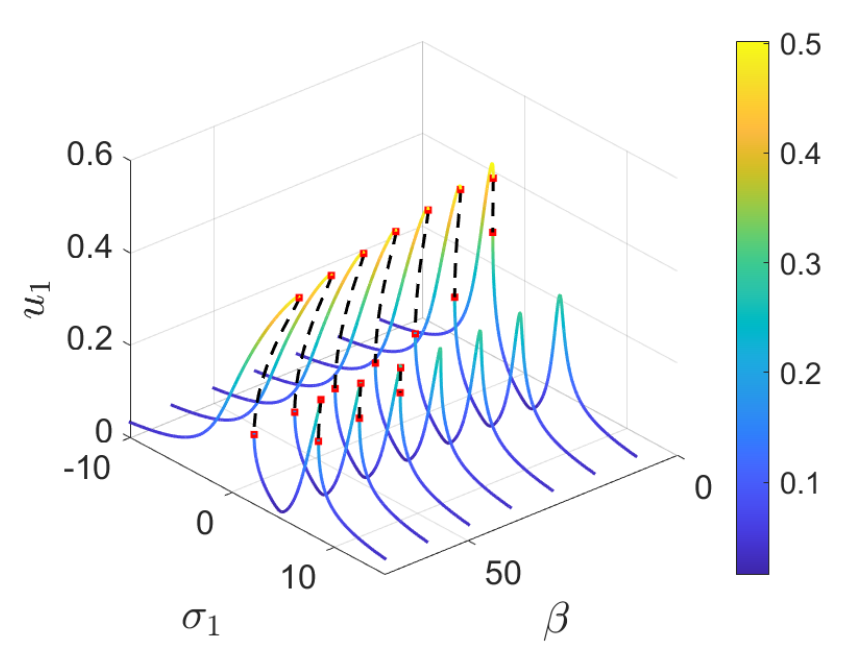}}
  \subfigure[]{\includegraphics[scale=.4]{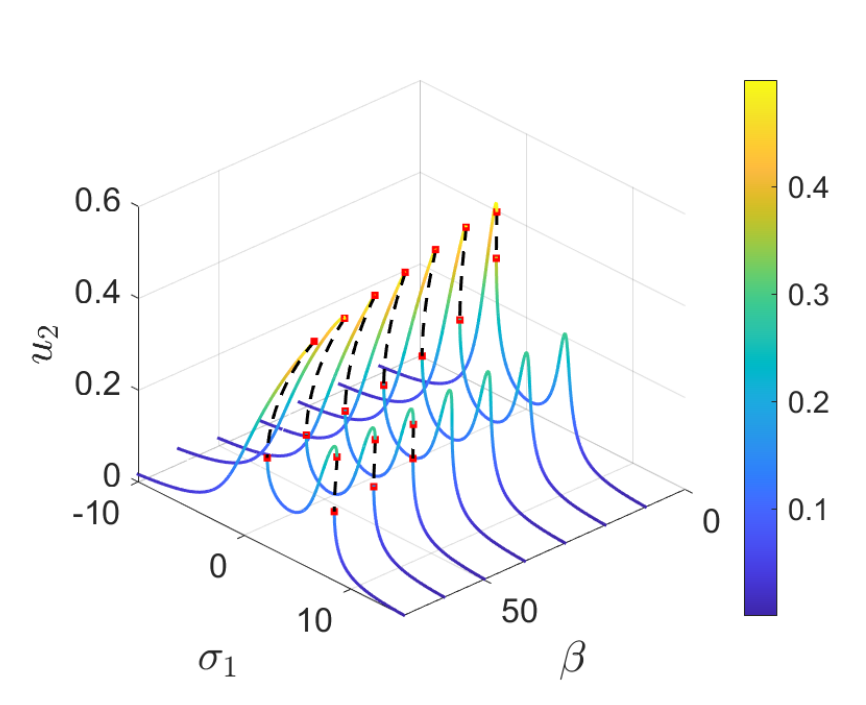}}
\caption{Frequency response of (a) oscillator $X$ and (b) oscillator $Y$ versus nonlinear coefficient $\beta$ and detuning parameter $\sigma_{1}$. In this case, $f = 1.0$, $d = 1.0$, $\epsilon = 0.1$, $\sigma_2 = 7.3$, $\delta = 1.0$, and $\beta$ ranges from $10$ to $70$.}
\label{fig:3d_beta}
\end{figure}

In this section, Fig. \ref{fig:3d_delta} demonstrates a series of frequency responses with varying coupling coefficient $\delta$ ranging from $0.2$ to $2.0$. It is evident that as $\delta$ increases, the second peak drifts away from the first. As a result, the effective characteristics of the frequency response, such as the \emph{distance between the two peaks}, are crucial in estimating the coupling coefficient. Fig. \ref{fig:3d_beta} depicts the frequency response curves for varying nonlinear coefficient $\beta$ ranging from $10$ to $70$. A potential feature to identify and measure the nonlinear coefficient in Fig. \ref{fig:3d_beta} is the changing \emph{curvature} of the resonance peaks.

\begin{figure}[!ht]
\centering
  \subfigure[]{\includegraphics[scale=.4]{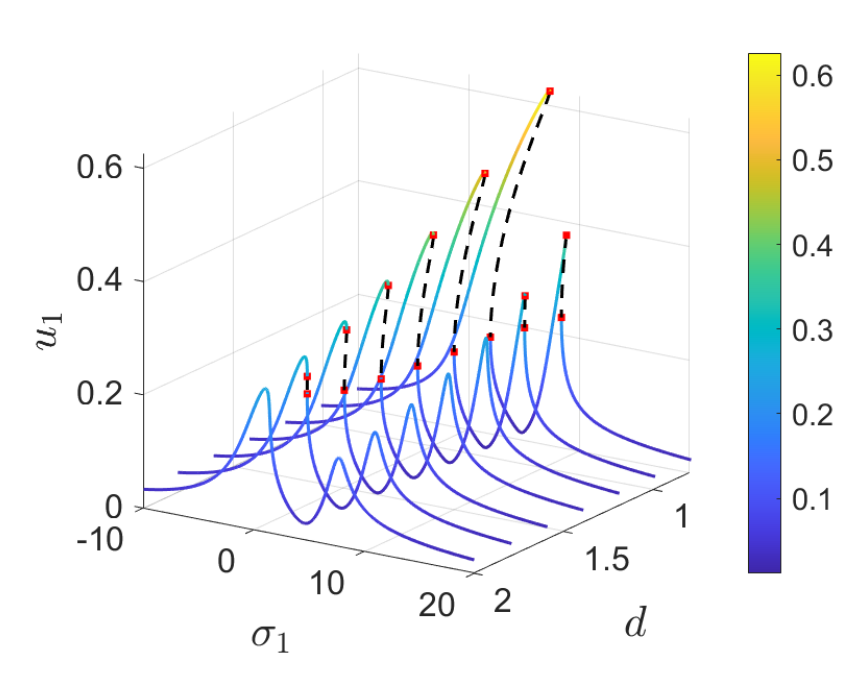}}
  \subfigure[]{\includegraphics[scale=.4]{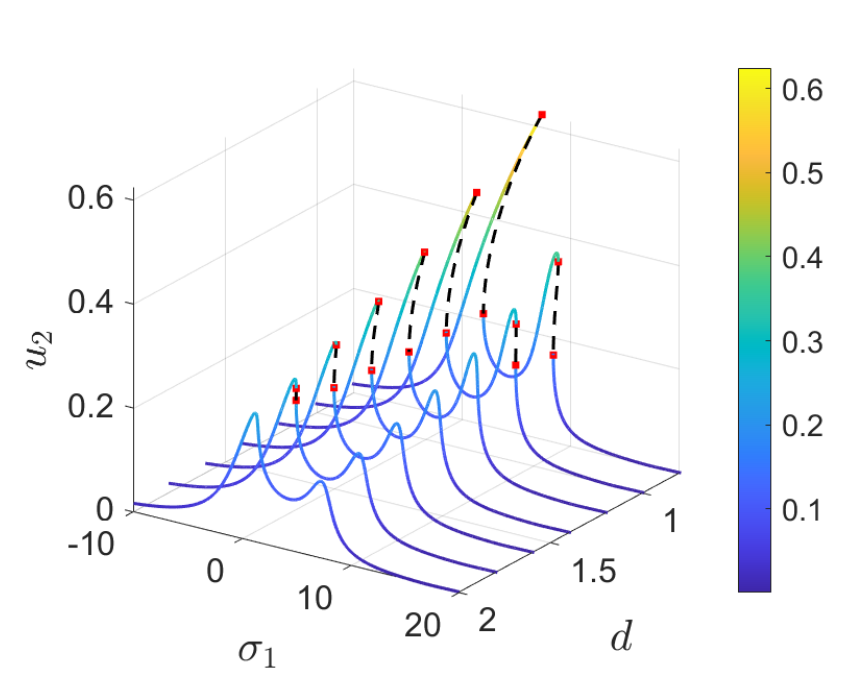}}
\caption{Frequency response of (a) oscillator $X$ and (b) oscillator $Y$ versus damping $d$ and detuning parameter $\sigma_{1}$. In this case, $d$ changes from $0.8$ to $2.0$, $f = 1.0$, $\epsilon = 0.1$, $\sigma_2 = 7.3$, $\delta = 1.0$, and $\beta=50$.}
\label{fig:3d_d}
\end{figure}

\begin{figure}[!ht]
\centering
  \subfigure[]{\includegraphics[scale=.4]{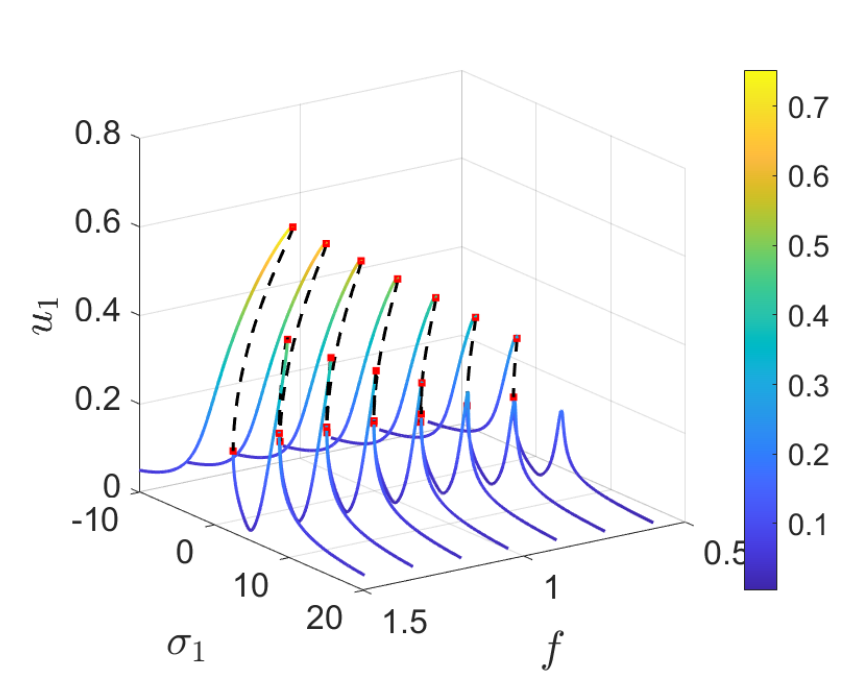}}
  \subfigure[]{\includegraphics[scale=.4]{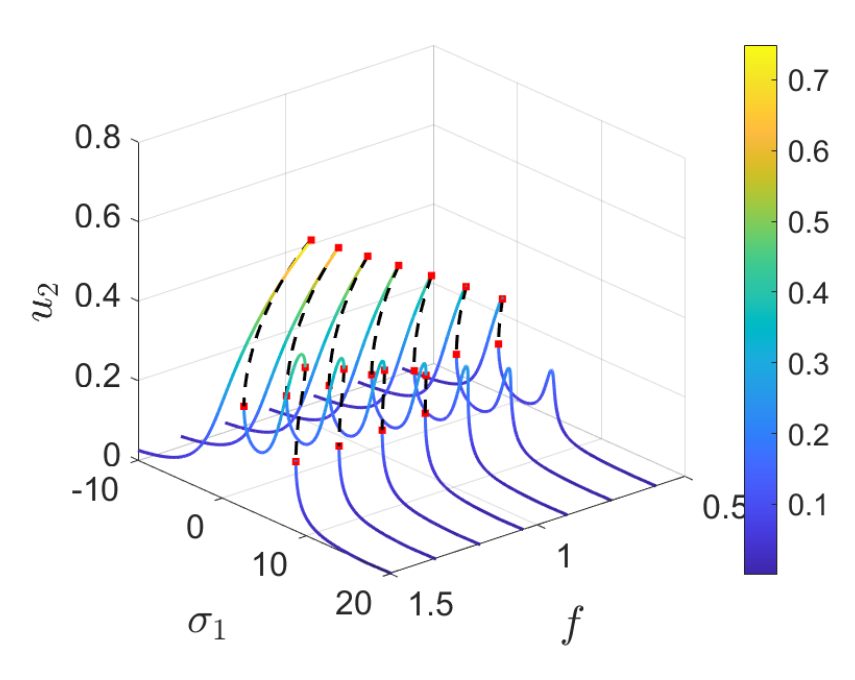}}
\caption{Frequency response of (a) oscillator $X$ and (b) oscillator $Y$ versus excitation amplitude $f$ and detuning parameter $\sigma_{1}$. For these plots, $f$ ranges from $0.5$ to $1.5$, $d=1.0$, $\epsilon = 0.1$, $\sigma_2 = 7.3$, $\delta = 1.0$, and $\beta=50$.}
\label{fig:3d_f}
\end{figure}

The effect of damping on frequency response is shown in Fig.~\ref{fig:3d_d} for damping coefficient $d$ ranging from $0.8$ to $2.0$. Fig.~\ref{fig:3d_f} depicts a series of frequency response curves corresponding to different amplitudes of excitation $f$ ranging from $0.5$ to $1.5$. It can be observed from Figs.~\ref{fig:3d_d} and~\ref{fig:3d_f} that the amplitudes and frequencies at the resonant peaks change with the variation in damping and amplitude of excitation, and therefore serve as potential features to estimate damping and amplitude of excitation.

\section{Data-based modeling}
\label{se:data_based modeling}
\subsection{Feature extraction}
\label{se:feature_extraction}
\begin{figure}[!ht]
\centering
  \subfigure[]{\includegraphics[scale=.3]{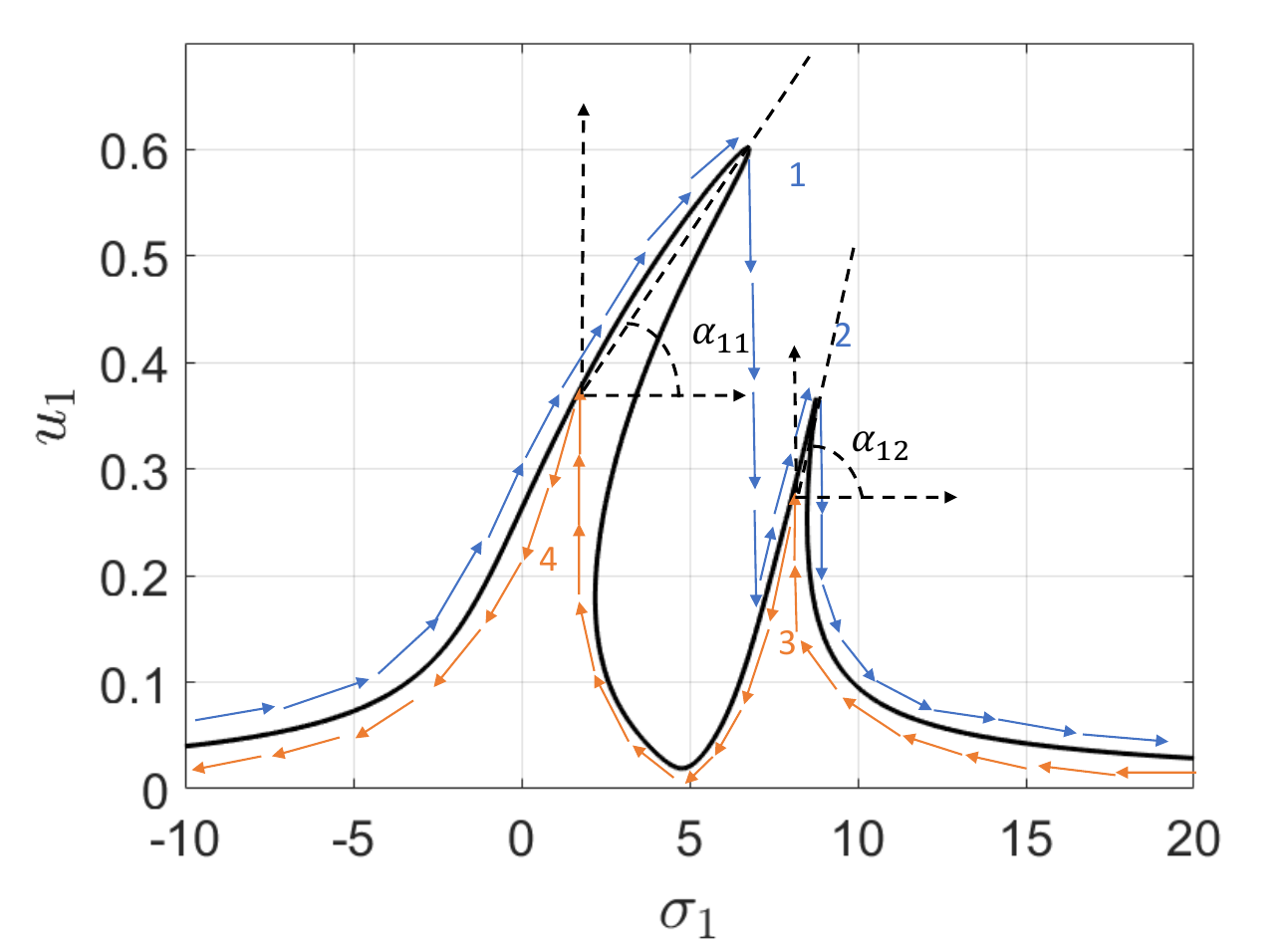}}
  \subfigure[]{\includegraphics[scale=.3]{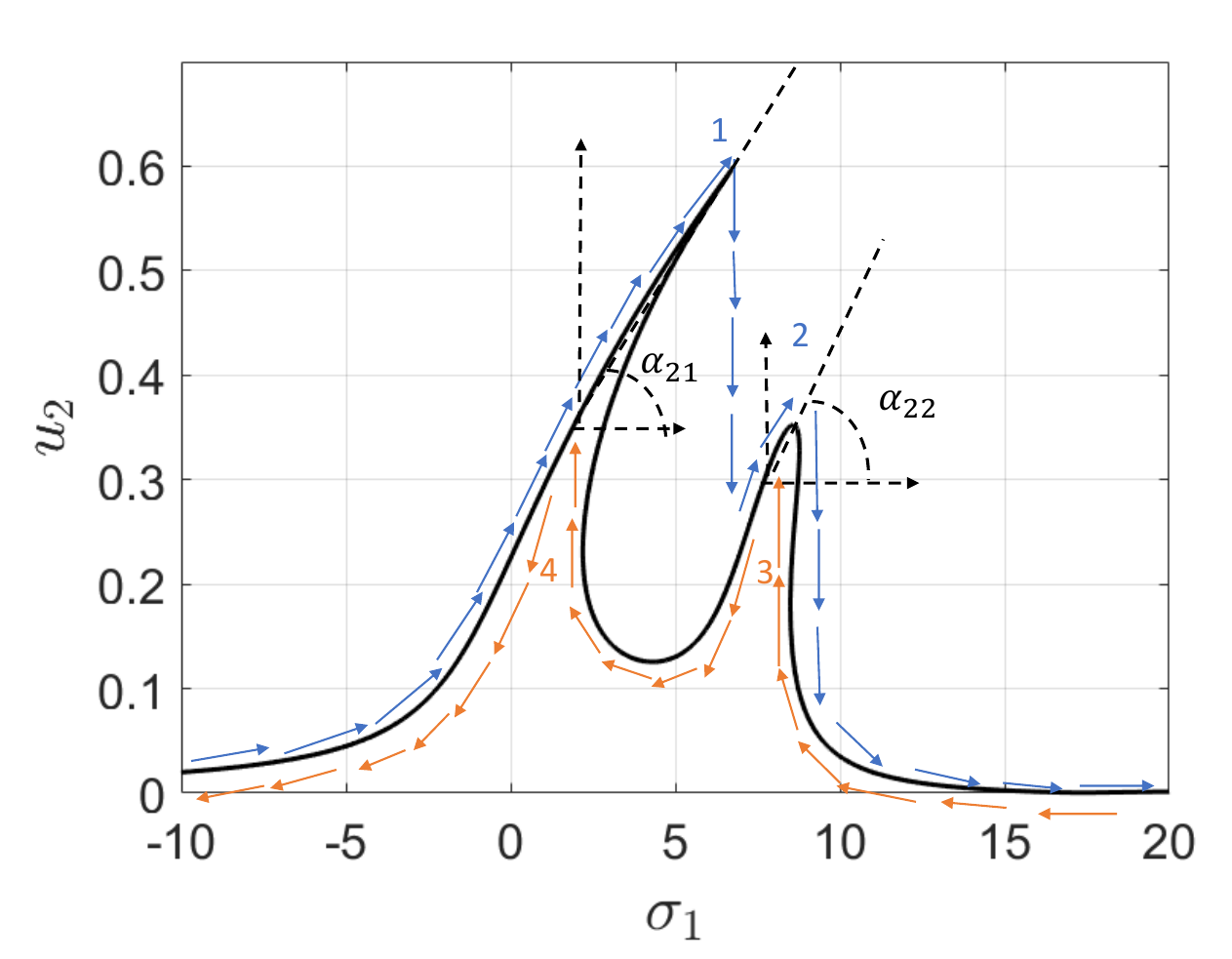}}
\caption{Feature extraction: (a) oscillator $X$, (b) oscillator $Y$}
\label{fig:features}
\end{figure}

The asymptotic solutions are leveraged to map changes in parameters onto geometrical changes in the frequency response by feature extraction. Fig.~\ref{fig:features} depicts the feature extraction from the nonlinear frequency responses of $X$ and $Y$. The jump phenomenon observed in the nonlinear frequency response indicates the occurrence of bifurcation and the switching of stable states. These jump points are the most pronounced and sensitive characteristics of the parameter changes. We label these points by numbers 1 and 2 for jump-down and 3 and 4 for jump-up in Fig.~\ref{fig:features}. We select the frequencies and amplitudes at these points as our features and denote them by $f_{ij}$, $p_{ij}$ where $i=1,2$ corresponds to oscillators $X$ and $Y$, and $j=1,2$ and $3,4$ corresponds to jump-down and jump-up points. Furthermore, the bending of the frequency response curve determined by the slopes at bifurcation points is a quantitative measure of the nonlinearity of the system. As a result, the slopes $\alpha_{ij}$ depicted in Fig. \ref{fig:features} are chosen as features, where $i=1,2$ corresponds to oscillator $X$ and $Y$ and $j=1,2$ denote the bending at the first resonance and the second resonance.

\begin{figure}[!ht]
\centering
\subfigure[]{\includegraphics[scale=0.45]{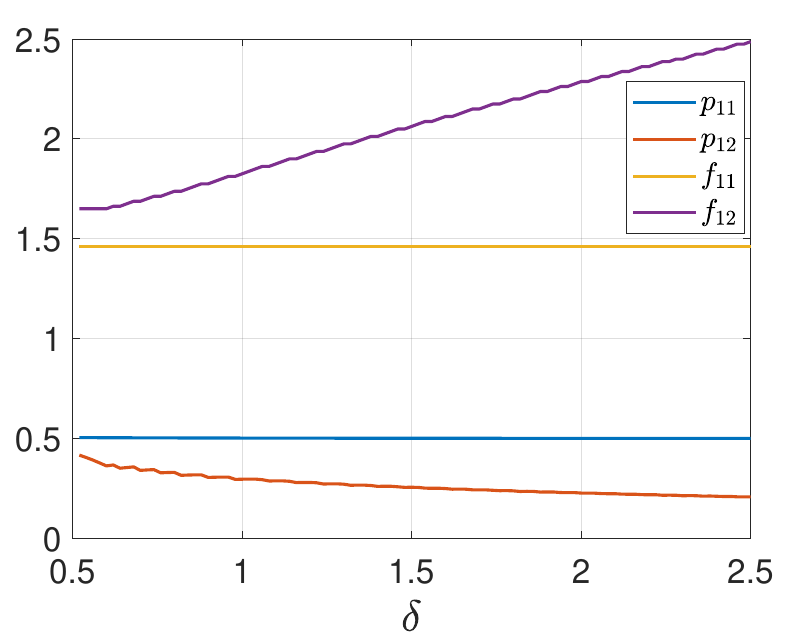}}
\subfigure[]{\includegraphics[scale=0.45]{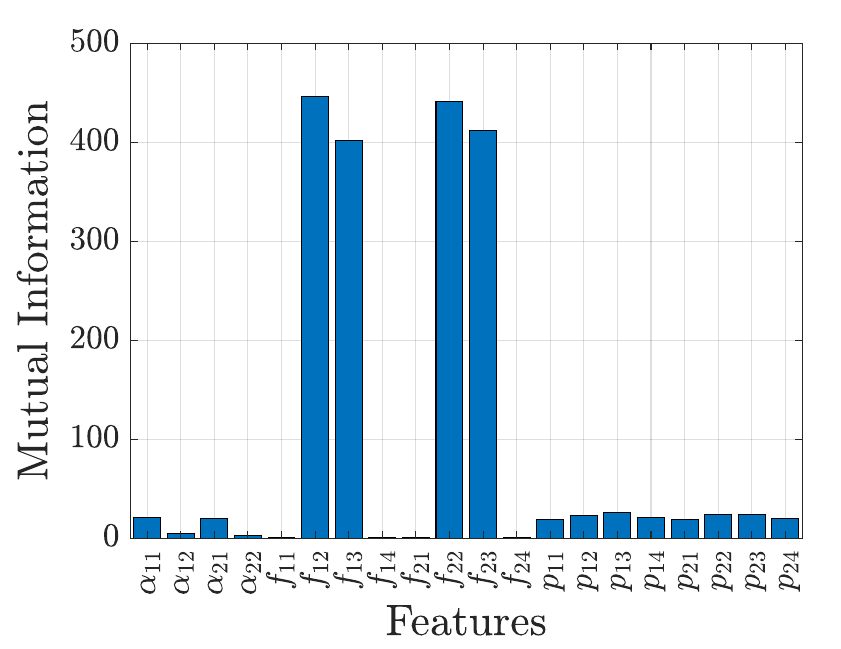}}
  \caption{ (a) Variation of features with respect to coupling coefficient, (b) mutual information between physics-based features and predicted coupling coefficient $\delta$}
\label{fig:feature_example}
\end{figure}

In summary, the feature space includes a total of twenty features, ten features for oscillator $X$ and ten features for oscillator $Y$. Fig. \ref{fig:feature_example} (a) demonstrates several features versus the coupling coefficient $\delta$. We can observe relatively linear relationships between the features $p_{12}$ and $f_{12}$ versus the coupling coefficient $\delta$, indicating that these features are promising to track variations of $\delta$. Also, features $p_{11}$ and $f_{11}$ which are apparently invariant to variations in $\delta$ may not be useful in this situation. It is worth noting that our feature space may hence contain superfluous features, i.e., they may not have sufficient information to contribute to the prediction of the output of the machine learning network. Hence, we need to weed them out and create an optimal feature set. Further, these `optimal' feature sets would very likely be different for different scenarios (as explained below). In order to accomplish this important task for ranking and picking the features that are most likely to provide the best predictions (that we will call `optimal'), we use the concept of mutual information (MI). MI will be used to rank the features and generate an effective feature set; it should be noted that although domain knowledge (and perhaps, intuition) from nonlinear dynamics was used to pick the \emph{initial} set of features, assessing the relative importance of features and finalizing the selection can be carried out in a systematic manner only by employing tools such as MI.

\subsection{Feature ranking with mutual information }
Based on preceding discussions, a qualitative analysis of the features is performed in the following section. The mutual information (MI) between features and varying parameters is calculated to estimate and rank features~\cite{kappaganthu2011feature,Kappaganthu2011,wang2014feature} as explained below.

Let $x$ and $y$ be random variables, which are subject to the Probability Density Function (PDF) $p(x)$ and $p(y)$. The entropy and mutual information can be defined as \eq{eq:entropy} and \eq{eq:Define_MI}.
\begin{equation}
H(x) = -\int p(x) \log p(x)dx
    \label{eq:entropy}
\end{equation}

\begin{equation}
I(x,y) = -\int p(x,y) \log \frac{p(x,y)}{p(x)p(y)}dxdy
    \label{eq:Define_MI}
\end{equation}

Hence, it follows that
\begin{equation}
    I(x,y) = H(y) - H(y|x)
\end{equation}
Let $x_{\rm{i}}$ be the random variable with PDF $p(x_{\rm{i}})$ corresponding to the $i^{\rm{th}}$ feature and $y_{\rm{k}}$ be the $k^{\rm{th}}$ coefficient with PDF $p(y_{\rm{k}})$. The entropy and mutual information of $i^{\rm{th}}$ coefficient and $k^{\rm{th}}$ feature are calculated by substituting $x_{\rm{i}}$ and $y_{\rm{k}}$ into \eq{eq:entropy} and \eq{eq:Define_MI}. Thus, $p(y_{\rm{k}})$ and $p(y_{\rm{k}}|x_{\rm{i}})$ are the prerequisites determined by the parameterization. According to the Bayesian rule, we obtain the following equations.
\begin{eqnarray}
\begin{aligned}
I(x_i,y) &= \sum_{k=1}^{N_k}p(y_k)\log p(y_k) \\& -\int \sum_{k=1}^{N_k} p(x_i|y_k)p(y_k)\log p(x_i|y_k)
    \label{eq:final_MI}
\end{aligned}
\end{eqnarray}

The MI of various features mentioned in Section \ref{se:feature_extraction} is obtained by \eq{eq:final_MI} and is visualized in Fig.~\ref{fig:feature_example} (b). The higher beams denote more MI between features and varying coefficients (coupling coefficient $\delta$ in this case). Feature $f_{12}$ has a larger mutual information than feature $f_{11}$, which validates our observation in Fig.~\ref{fig:feature_example} (a).

\subsection{Development of Artificial Neural Network (ANN)}

In this section, we develop an ANN for parameter identification using the ranked features. Conventional techniques like grid search and cross-validation are utilized to determine the hyperparameters~\cite{simon2009neural}. A two-layer neural network with a single hidden layer is selected to map such features onto different coupling coefficients $\delta$. In the training phase, the regularization technique is employed to prevent over-fitting. The candidate construction of the neural network is leveraged to search for the best size and appropriate feature sets from all features. The search process starts from the best features and proceeds to the worst features estimated by the MI. Firstly, we feed the top one feature into the neural network based on the mutual information presented in Fig. \ref{fig:feature_example} (b), then we use the grid search algorithm to find specific hyperparameters and record the lowest Root Mean Square Error (RMSE). The feature set is then formed using the top one and top two ranked features, and the first phase is repeated. We repeat this process until all of the features in our feature space are included. This process is called forward search.

\begin{figure}[!ht]
\centering
\subfigure[]{\includegraphics[scale=0.45]{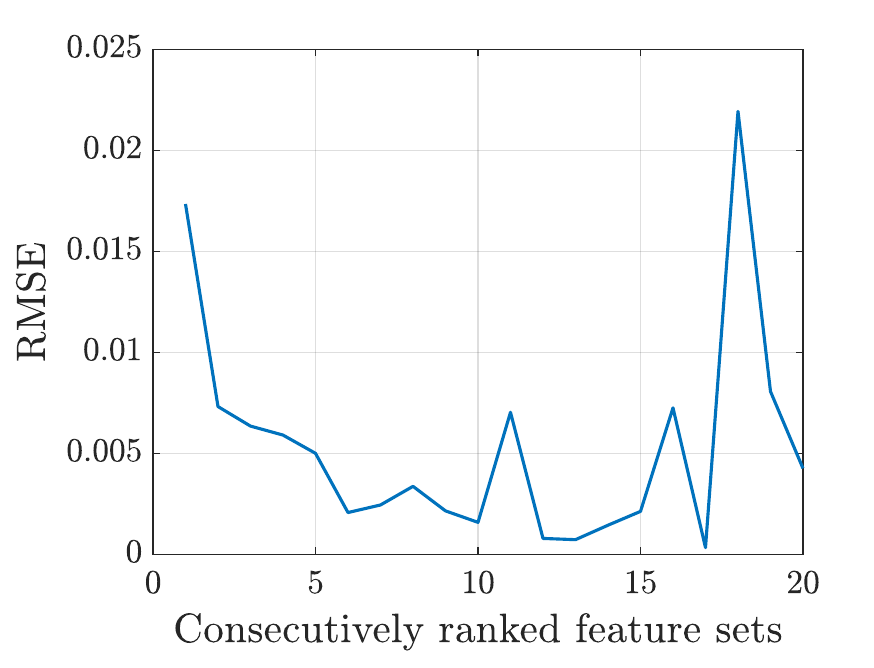}}
\subfigure[]{\includegraphics[scale=0.45]{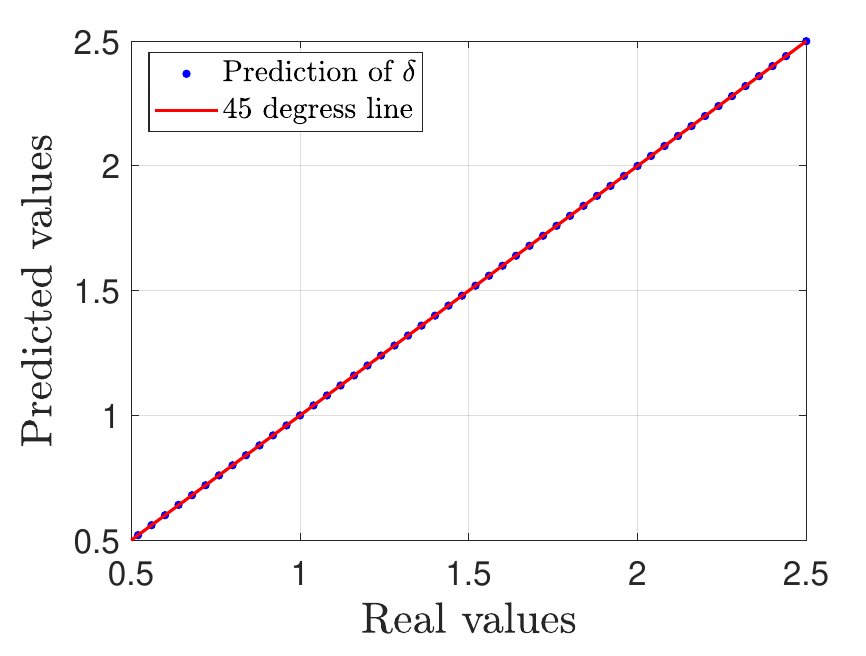}}
  \caption{(a) Forward search for feature set, (b) test results of predicting coupling coefficient $\delta$}
\label{fig:search}
\end{figure}

The observed RMSE versus consecutively ranked feature sets are shown in Fig. \ref{fig:search} (a) which is an important reference to build a feature set for identifying coefficients $\delta$. Note that the error decreases monotonically as features add up to the sixth feature. Further addition in features increases the error due to over-fitting as well as the low quality of the features. Moreover, in Fig. \ref{fig:feature_example} (b), it is observed that the dominant MI is for four features $f_{12}$, $f_{13}$, $f_{22}$ and $f_{23}$. This information is taken into account and we choose to extract four features $f_{12}$, $f_{13}$, $f_{22}$ and $f_{23}$ to build our feature set for coupling coefficient estimation after conducting numerous tests. We have chosen one hidden layer with 15 nodes for the construction of the ANN. One hundred sets of data with the four above-mentioned features in each set are generated. The training set contains 60\% of the data, while the validation and testing set each contains 20\% of the data. We generate extra 40 sets of data with different sources to test the trained neural network. Test result with a final RMSE of $4.5\times 10^{-8}$ is shown in Fig. \ref{fig:search} (b), where The 45-degree line indicates the actual values (red line) and the blue points indicate the estimated value.

\section{Parameter estimation using adaptive modeling}
\label{se:simulation and comparison}

In the preceding section, an example under ideal conditions is used to interpret the proposed method. In reality, the variation of only one parameter is not guaranteed. In addition, system loads might be unpredictable at times. These aspects must be taken into account. Furthermore, we are interested not only in the coupling coefficient but also in damping which is very likely to change in real systems due to wear and contamination. In this section, we apply hybrid adaptive modeling to track both the coupling coefficient and damping coefficient simultaneously with random loads and different operational conditions. The gray box model, a conventional numerical method, is employed as a comparison.

\subsection{Estimation of coupling coefficient}
\label{sec:delta}
In this case, the ANN is trained for estimating the coupling coefficient using data obtained by parameterizing the system in \eq{eq:modulation2}. Each data set consists of a frequency response curve generated by varying the excitation amplitude $f$ randomly from $0.9$ to $1.1$, varying coupling coefficient $\delta$ from $1.0$ to $2.0$ and with a constant nonlinear coefficient $\beta$ set to $40$. The targeted parameter, i.e., the coupling coefficient, is estimated considering six different damping scenarios detailed in Table.~\ref{table:delta}. 

\begin{figure}[!ht]
\centering
  \subfigure[]{\includegraphics[scale=.3]{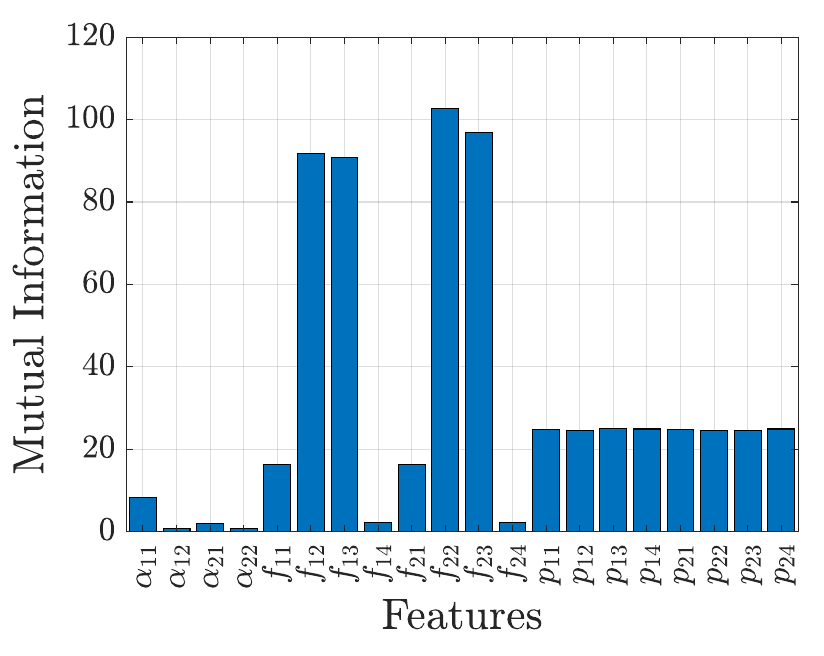}}
  \subfigure[]{\includegraphics[scale=.3]{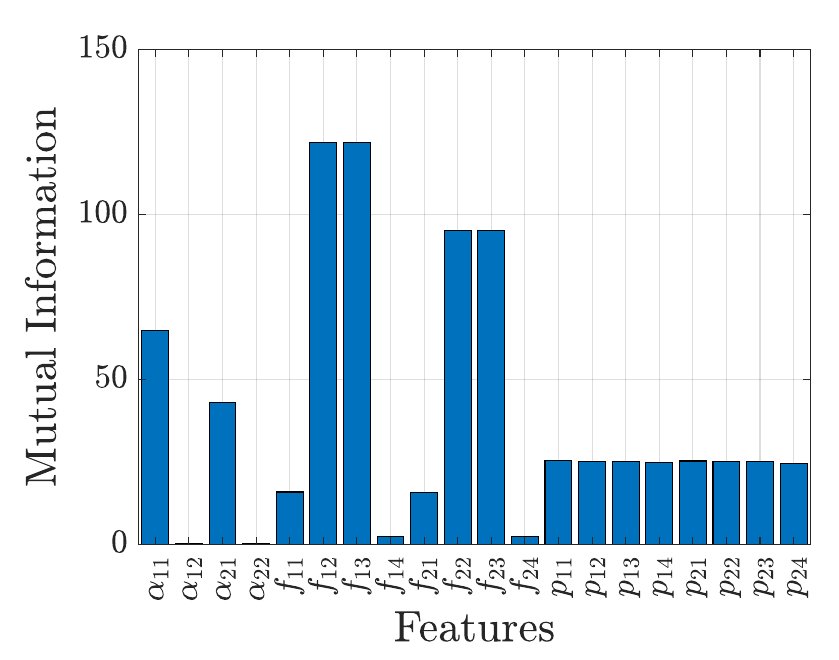}}
  \subfigure[]{\includegraphics[scale=.3]{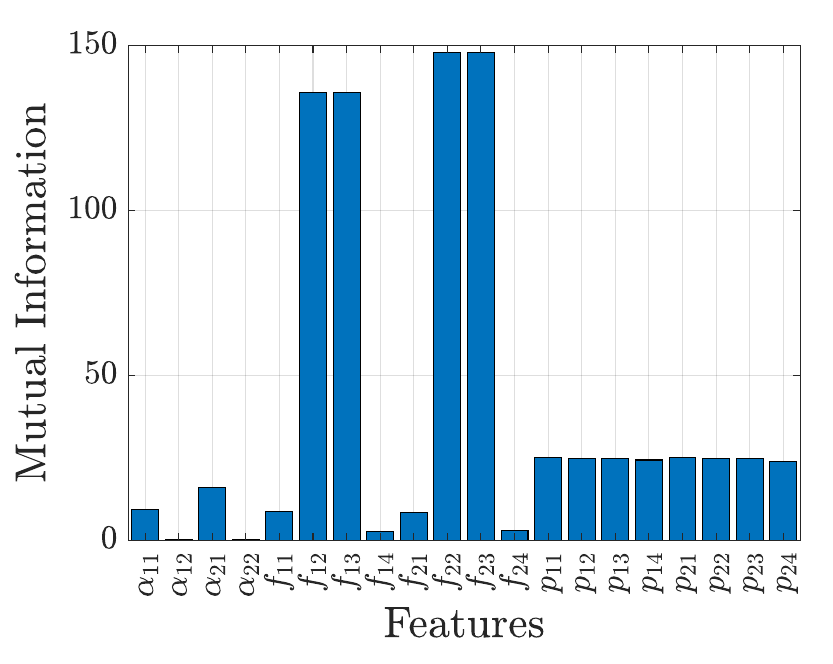}}
  \subfigure[]{\includegraphics[scale=.3]{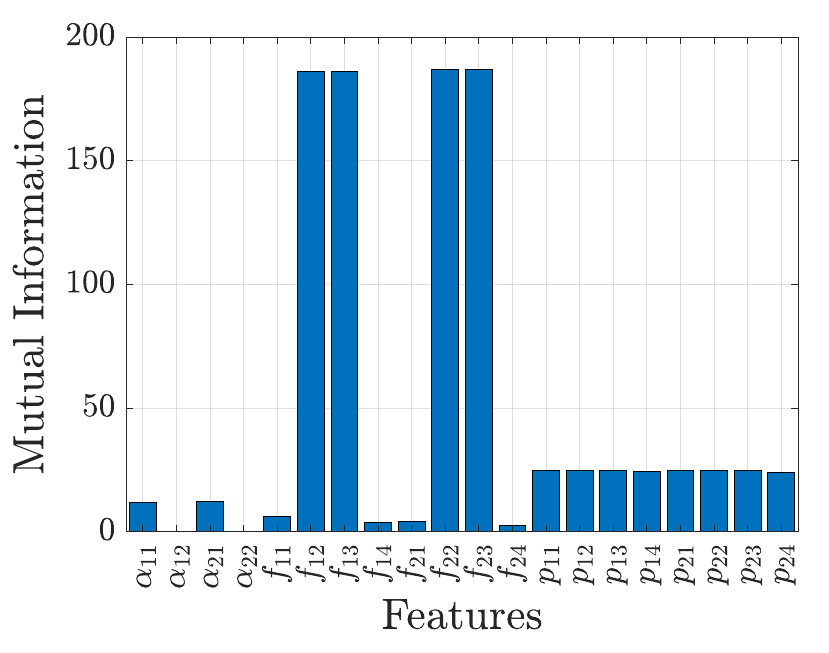}}
  \subfigure[]{\includegraphics[scale=.3]{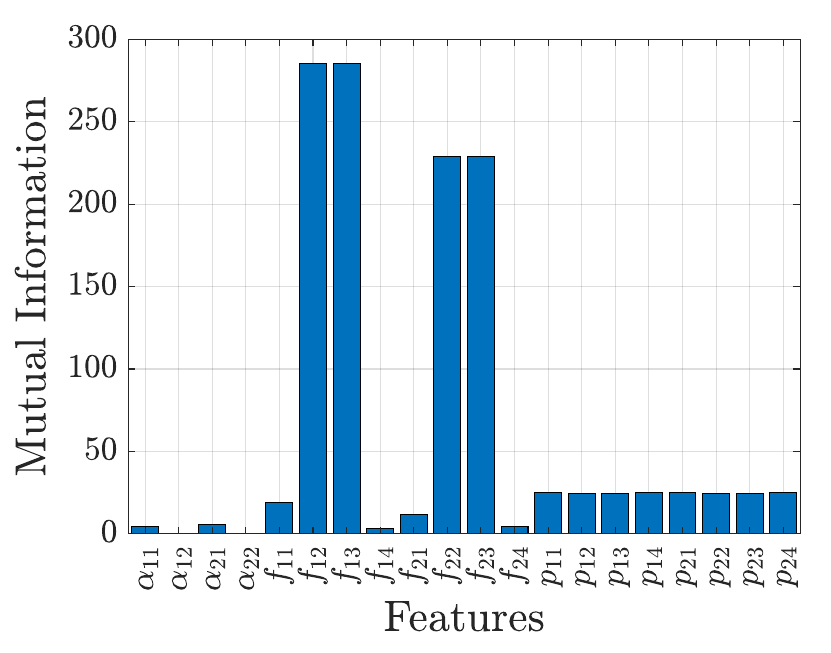}}
  \subfigure[]{\includegraphics[scale=.3]{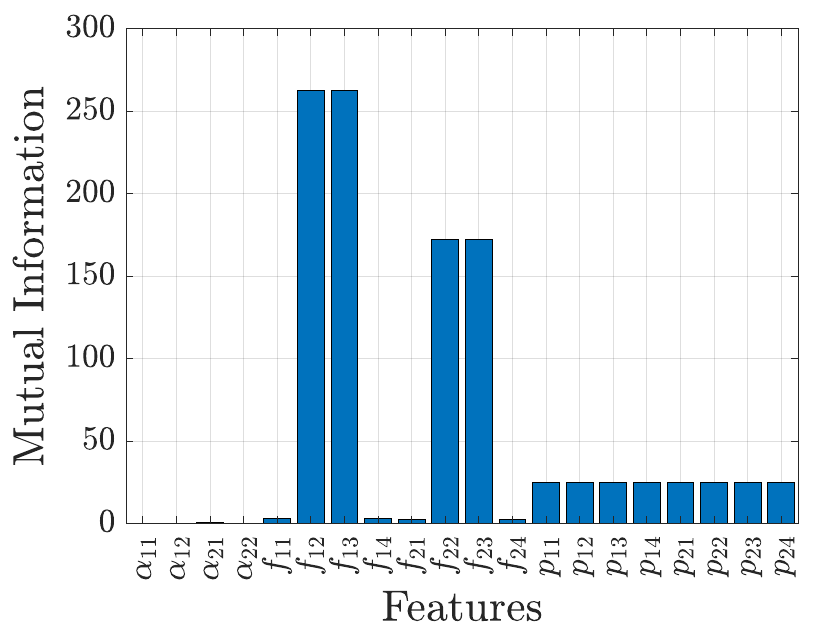}}
\caption{Feature ranking for predicting coupling coefficient $\delta$ using MI: (a) $d = 1.0$, (b) $d = 1.2$, (c) $d = 1.4$, (d) $d = 1.6$, (e) $d = 1.8$, (f) $d = 2.0$}
\label{fig:rank_delta}
\end{figure}

For each scenario, a hundred sets of feature spaces are generated as training sets, and the other thirty sets are generated as testing sets. The MI is calculated, and all features are ranked as shown in Fig. \ref{fig:rank_delta}. The absolute dominance of $f_{12}$, $f_{13}$, $f_{22}$ and $f_{23}$ under all damping conditions is clearly visible. It is also observed that as damping increases, the MI of amplitude-related features decreases. \ZL{A plausible explanation for various dominant features is as follows. The changing damping affects the system’s total energy. The jump phenomena are associated with transitions from one state of equilibrium to another, which have different levels of energy. Therefore, variations in damping impact the quality of features related to these jump phenomena}

\begin{figure}[!ht]
\centering
  \subfigure[]{\includegraphics[scale=.3]{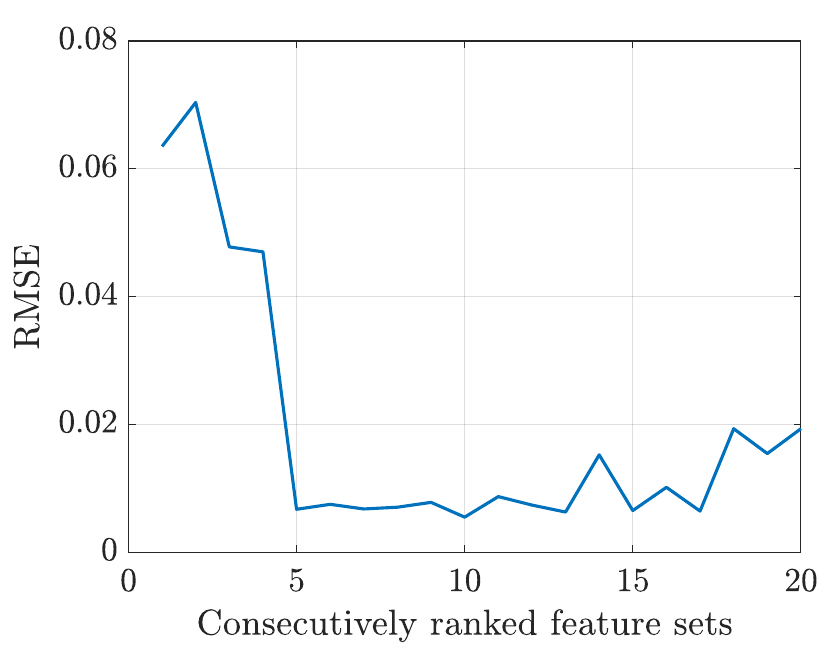}}
  \subfigure[]{\includegraphics[scale=.3]{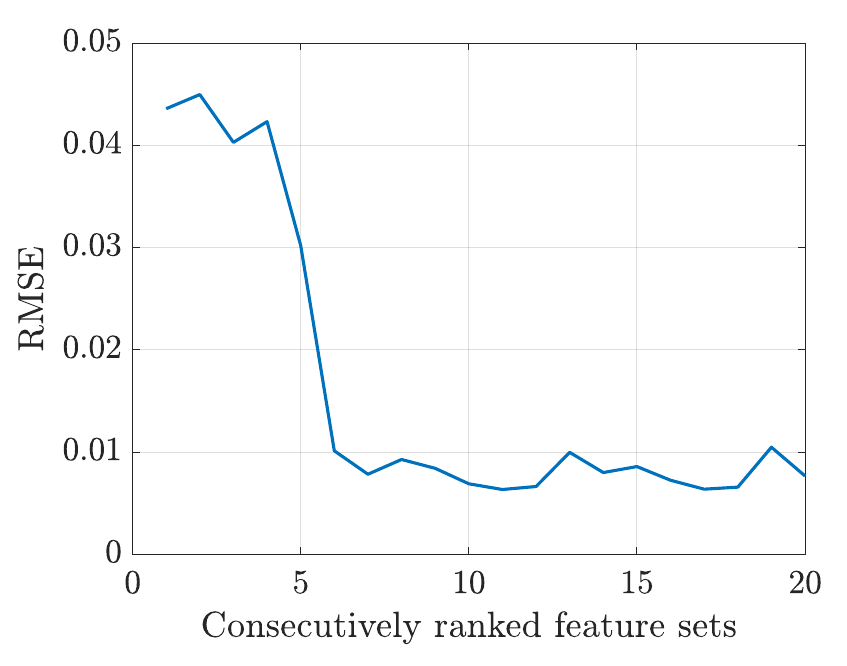}}
  \subfigure[]{\includegraphics[scale=.3]{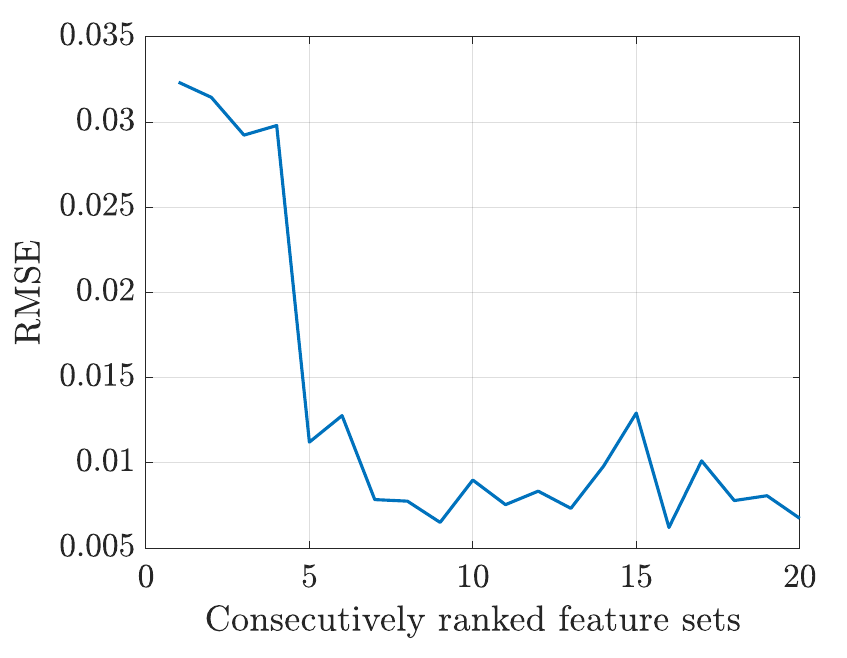}}
  \subfigure[]{\includegraphics[scale=.3]{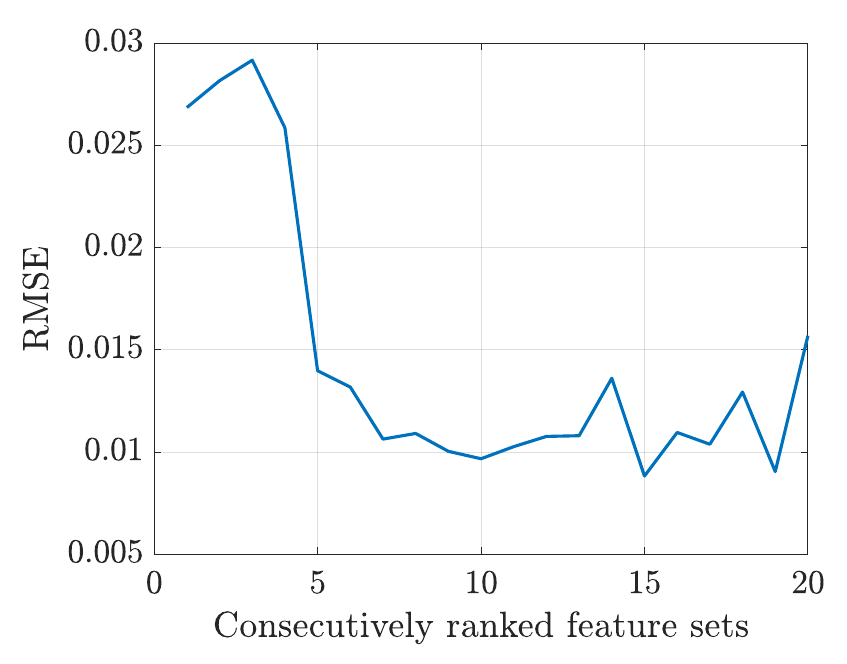}}
  \subfigure{\includegraphics[scale=.3]{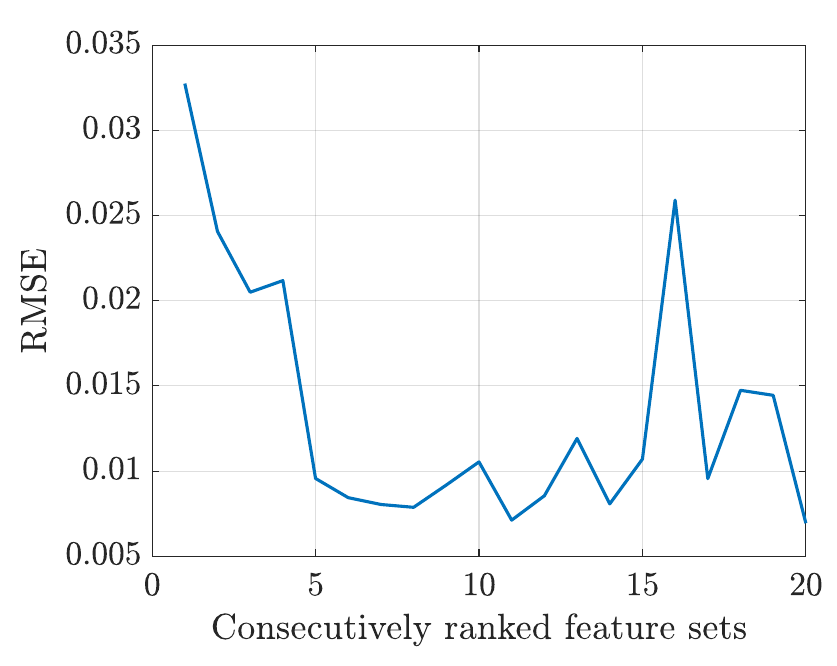}}
  \subfigure[]{\includegraphics[scale=.3]{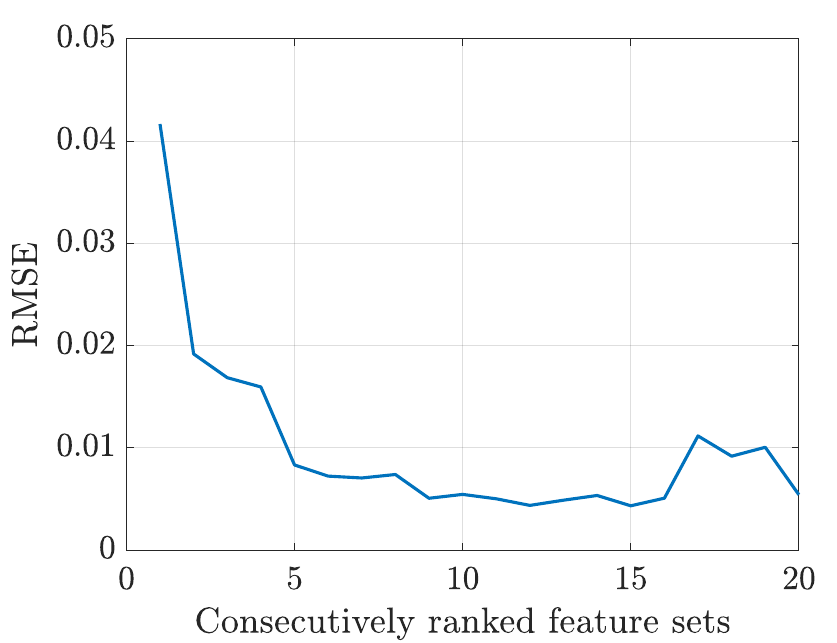}}
\caption{Forward search for predicting coupling coefficient $\delta$ using ranked features: (a) $d = 1.0$, (b) $d = 1.2$, (c) $d = 1.4$, (d) $d = 1.6$, (e) $d = 1.8$, (f) $d = 2.0$}
\label{fig:search_delta}
\end{figure}

The forward search results for the six scenarios are shown in Fig.~\ref{fig:search_delta}.
We can see a significant decline in RMSE value for a set of consecutive top-ranked features ranging from one to five for all situations; however, according to Fig.~\ref{fig:rank_delta}, the MI of the fifth consecutive top-ranked feature set changes for different damping values, so it is not robust enough. Meanwhile, the four features $f_{12}$, $f_{13}$, $f_{22}$, and $f_{23}$ have consistent advantages in MI over other features for all six scenarios. In other words, these features are sensitive to changes in $\delta$ but are yet resistant to damping variations. So we still choose the top four features as our feature set to train the neural network for tracking the changing coupling coefficients. Subsequently, repeating the above procedure, the test data is generated from other sources and is fed to the trained neural network to test the predictions of the coupling coefficient $\delta$.

\begin{figure}[!ht]
\centering
  \subfigure[]{\includegraphics[scale=.3]{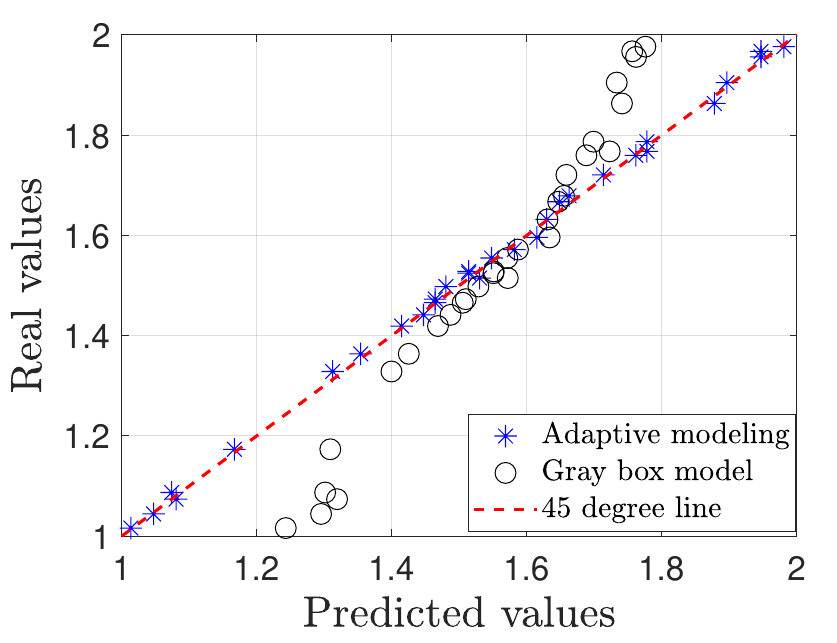}}
  \subfigure[]{\includegraphics[scale=.3]{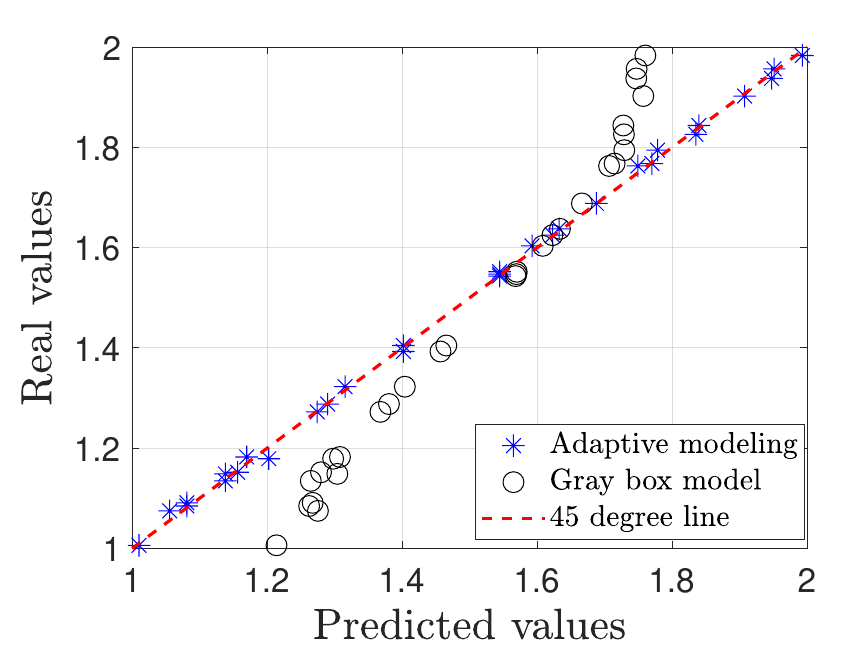}}
  \subfigure[]{\includegraphics[scale=.3]{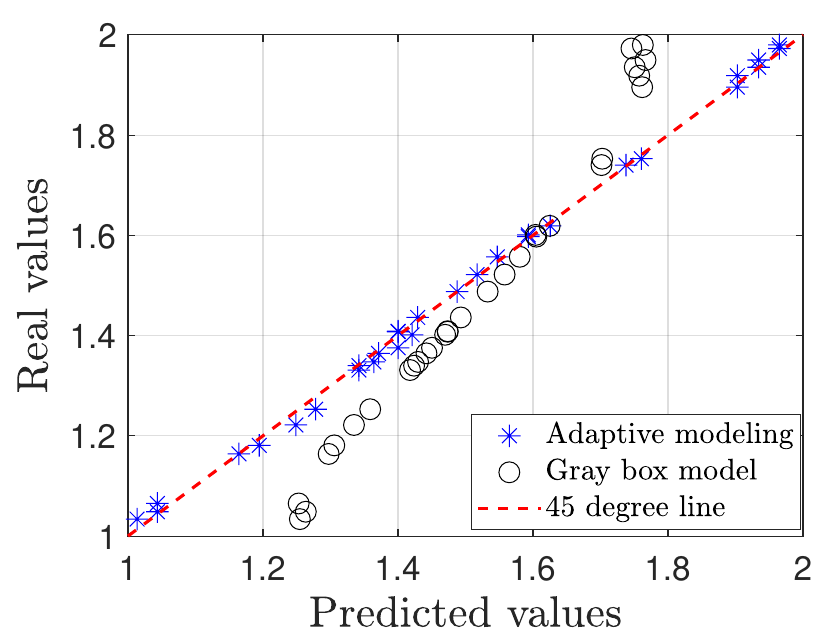}}
  \subfigure[]{\includegraphics[scale=.3]{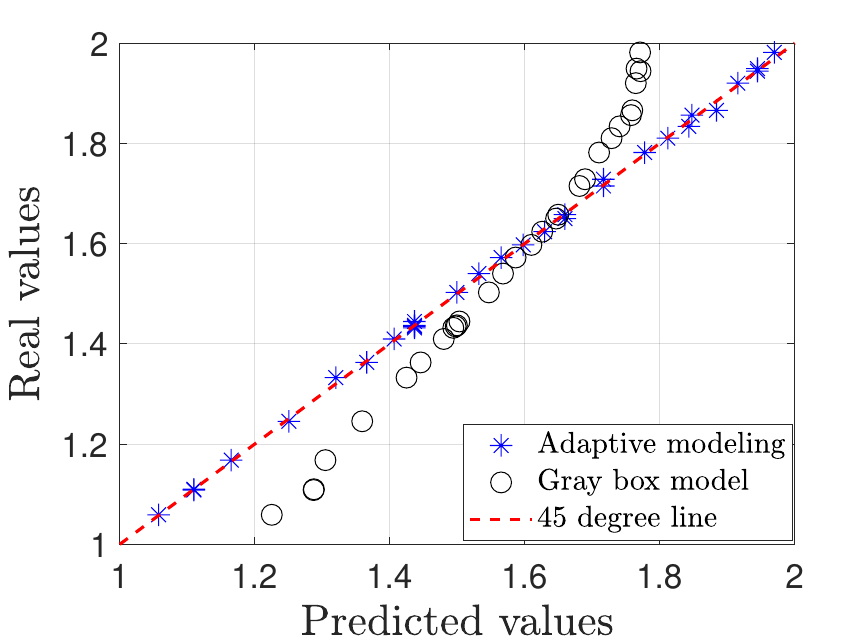}}
  \subfigure[]{\includegraphics[scale=.3]{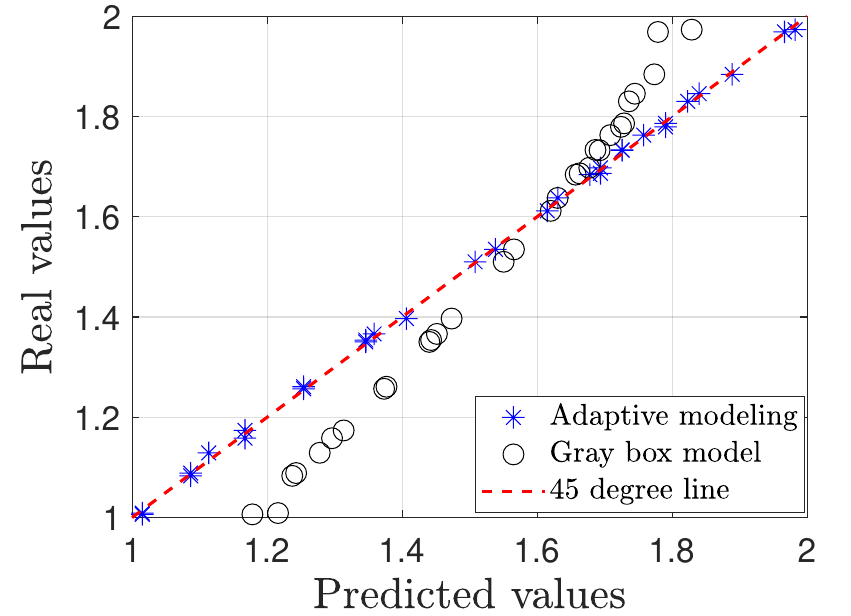}}
  \subfigure[]{\includegraphics[scale=.3]{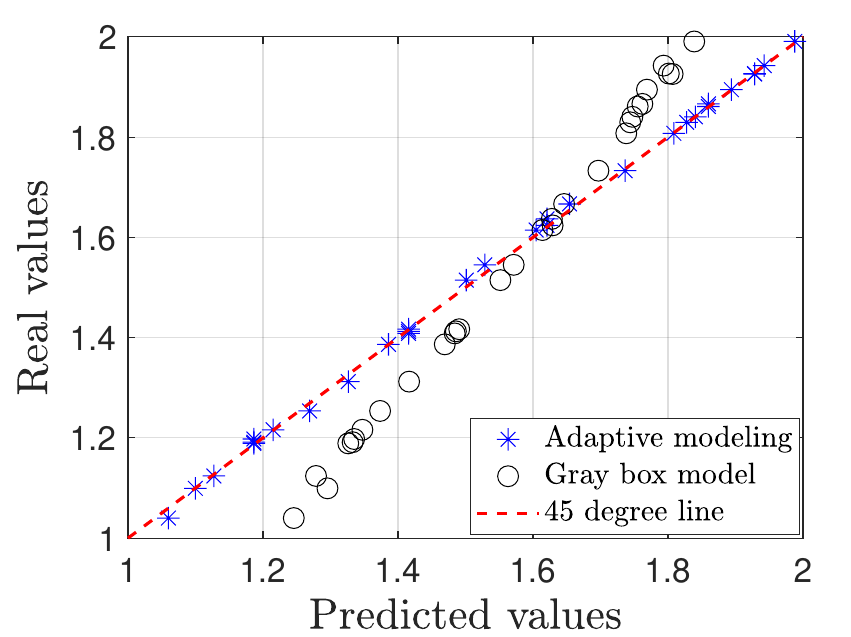}}
\caption{Comparison of adaptive model and gray box model for predicting coupling coefficients $\delta$: (a) $d=1.0$, (b) $d=1.2$, (c) $d=1.4$, (d) $d=1.6$, (e) $d=1.8$, (f) $d=2.0$}
\label{fig:Compare_delta}
\end{figure}

To demonstrate the effectiveness of the proposed technique under varied damping conditions, we compare the estimations from the present technique to the estimations from the widely used gray box model. Typically, a gray box model requires a time series as the input. Each time series data is generated for varying excitation $f$ and coupling coefficient $\delta$ and with a constant nonlinear stiffness. These data sets are then generated for six scenarios of damping as discussed above and the coupling coefficient $\delta$ is estimated. Finally, the estimations from the gray box model and the proposed adaptive modeling technique are compared and are shown in Fig.~\ref{fig:Compare_delta}.

The RMSE is used to estimate the performance of adaptive modeling and gray box modeling. For convenience, we use \ZL{$\mathrm{RMSE}_{a}$} to denote the RMSE of estimations from adaptive modeling, and \ZL{$\mathrm{RMSE}_{g}$} to represent the RMSE of estimations from gray box modeling. The results are listed in Table.~\ref{table:delta}. The hybrid adaptive model clearly outperforms the traditional gray box model in terms of accuracy. 

\ZL{From a computational efficiency standpoint, generating one hundred training sets with MATCONT required $2500$ seconds in total. The training process of the artificial neural network training process was even faster, completed in just $35$ seconds due to the structure of the shallow network and the use of only four input features. Once trained, the time needed for new predictions is negligible in a well-tuned model. On the other hand, it takes an average of $200$ seconds for the gray-box model to estimate a new parameter by feeding time series data with a specific excitation frequency. However, the choice of frequency highly impacts the parameter estimations due to parsimonious resonance and other dynamical phenomena with different excitation frequencies. To increase the accuracy, we take the average of ten estimations, corresponding to ten different frequencies, as the final result. There are $40$ test sets in this study which take more than $22$ hours by gray-box model with the time increasing significantly if the initial guess deviates larger from the true parameter. In comparison, the proposed method takes less than ten minutes with the trained model.}

\begin{table}[t]

\centering
\begin{tabular}{c | c | c | c}
\hline \hline
\textbf{Scenario} & \textbf{Damping} $d$ &  \ZL{$\mathrm{RMSE}_a$} & \ZL{$\mathrm{RMSE}_g$} \\
  \hline \hline
   1 & $1.0$ & $0.0112$ & $0.1224$\\ 
   \hline
   2 & $1.2$ &  $0.0092$ & $0.1227$\\
   \hline
   3 & $1.4$ & $0.0131$ & $0.1232$\\
   \hline
   4 & $1.6$ &
   $0.0068$ & $0.1061$\\ 	
   \hline
   5 & $1.8$ & $0.0108$  & $0.1134$\\
   \hline
   6 & $2.0$ & $0.0080$  & $0.1105$\\
   \hline
\end{tabular}
\caption{Performance comparison between gray box model and adaptive modeling for predicting coupling coefficient with different scenarios (\ZL{$\mathrm{RMSE}_a$}: the root mean square error of estimations from the hybrid adaptive modeling, \ZL{$\mathrm{RMSE}_g$}: the root mean square error of estimations from the gray box modeling)}
\label{table:delta}
\end{table}

\subsection{Estimation of damping}

Damping is another source of uncertainty in dynamic systems due to wear and other factors and hence it is essential to track the changing damping after a period of operation. In this section, the proposed approach is employed to estimate the changing damping. In this case, the  ANN is trained for estimating damping using data obtained by parameterizing the system in \eq{eq:modulation2}. Each data set consists of a frequency response curve generated by varying excitation amplitude $f$ randomly from $0.9$ to $1.1$, varying coupling coefficient $\delta$ from $1.0$ to $2.0$, varying damping coefficient $d$ from $1.0$ to $2.0$ and with a constant nonlinear coefficient $\beta$ set to $40$. The targeted parameter, i.e., the damping coefficient is estimated considering six different coupling coefficient scenarios detailed in Table.~\ref{table:damping}.

\begin{figure}[!ht]
\centering
  \subfigure[]{\includegraphics[scale=.3]{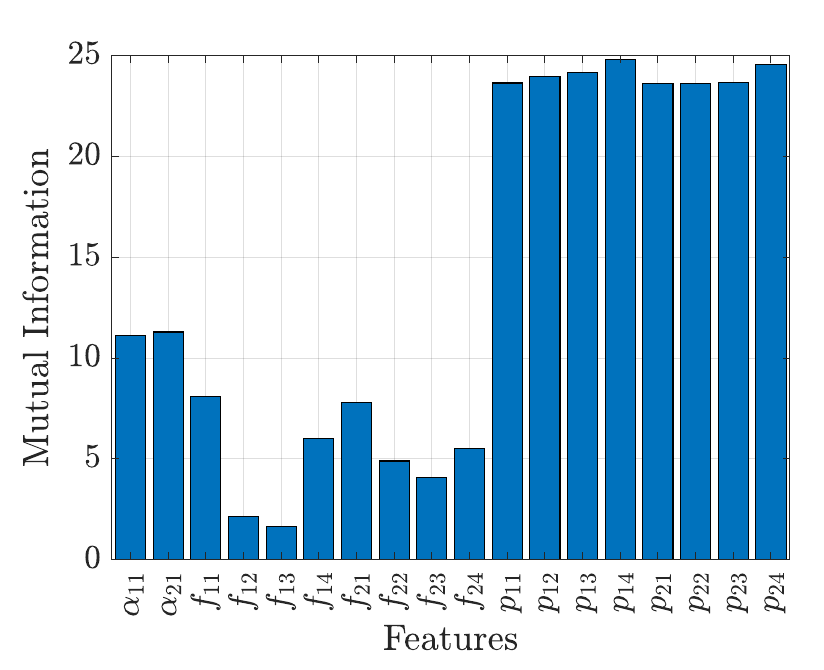}}
  \subfigure[]{\includegraphics[scale=.3]{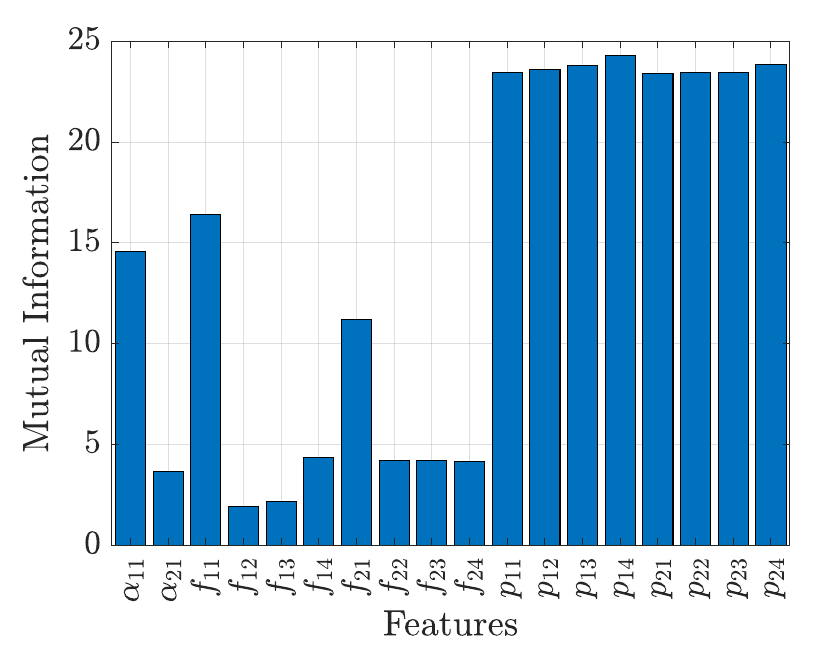}}
  \subfigure[]{\includegraphics[scale=.3]{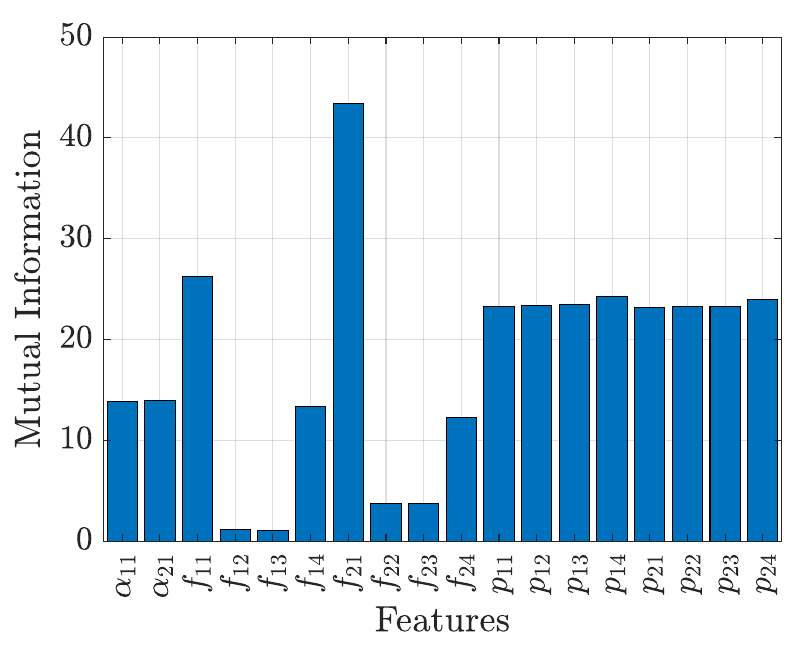}}
  \subfigure[]{\includegraphics[scale=.3]{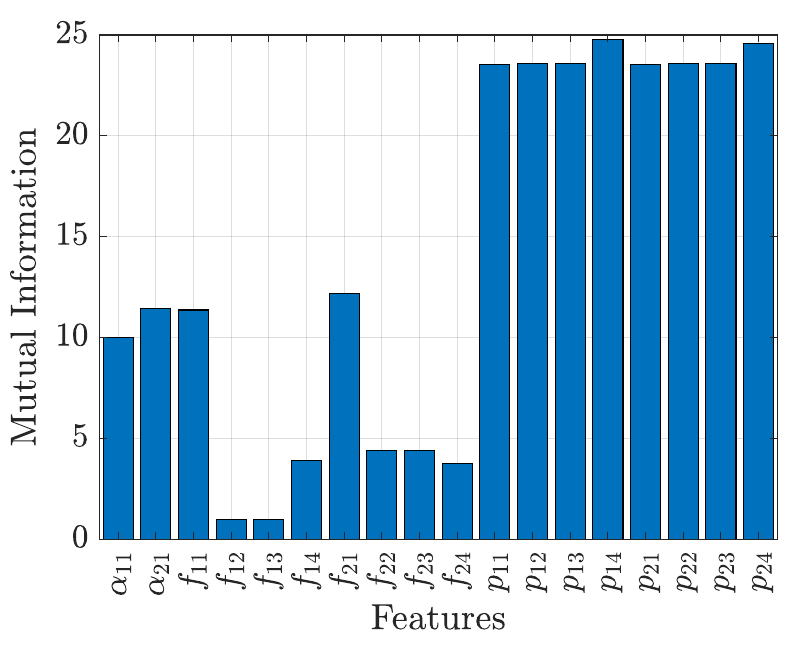}}
  \subfigure[]{\includegraphics[scale=.3]{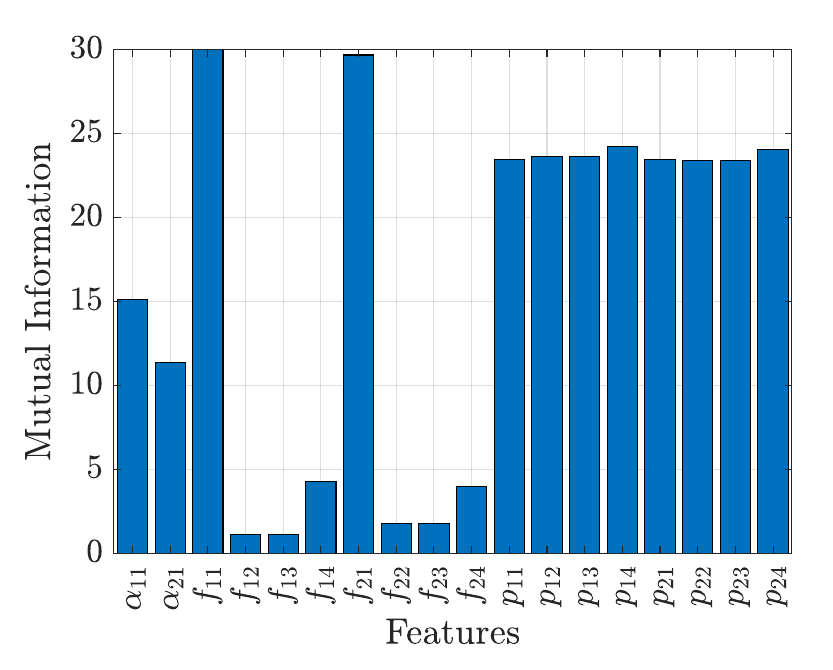}}
  \subfigure[]{\includegraphics[scale=.3]{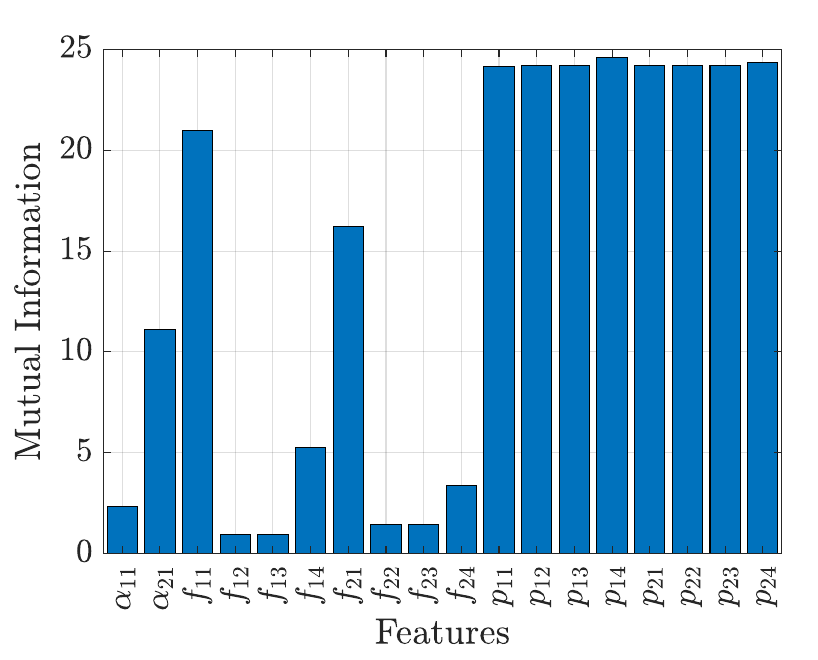}}
\caption{Feature ranking for predicting damping $d$ using MI: (a) $\delta = 1.0$, (b) $\delta = 1.2$, (c) $\delta = 1.4$, (d) $\delta = 1.6$, (e) $\delta = 1.8$, (f) $\delta = 2.0$}
\label{fig:rank_d}
\end{figure}

Similarly, a hundred sets of feature spaces are generated as training sets and the other thirty sets as testing sets for each scenario. In different scenarios, the dominating features resulting from feature ranking and selection (using MI) are different, thereby indicating that alterations in coupling coefficients and random loads yield diverse effects on these feature sets for estimating damping as shown in Fig.~\ref{fig:rank_d}. However, $f_{11}$ and $f_{21}$ which are superior to other frequency related frequency-related features, and $p_{14}$, $p_{24}$ which are the best among other amplitude-related features, are consistent for all the scenarios. \ZL{In this system, amplitudes are highly dependent on damping, resulting in amplitude-related features demonstrating higher mutual information. In each scenario, external forces are set at random values to intentionally introduce uncertainty, which significantly impacts the amplitudes and leads to variations in the dominant features.}

\begin{figure}[!ht]
\centering
  \subfigure[]{\includegraphics[scale=.3]{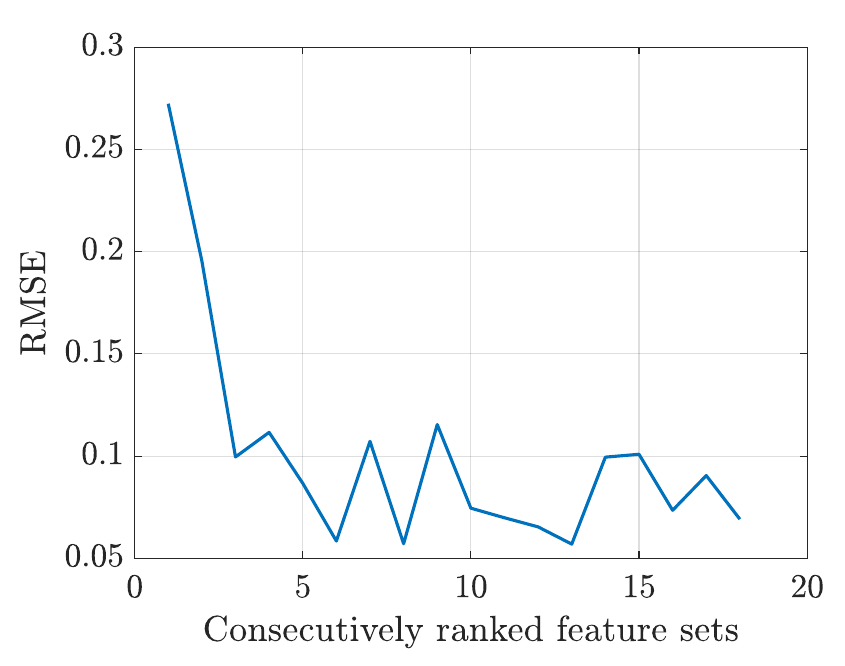}}
  \subfigure[]{\includegraphics[scale=.3]{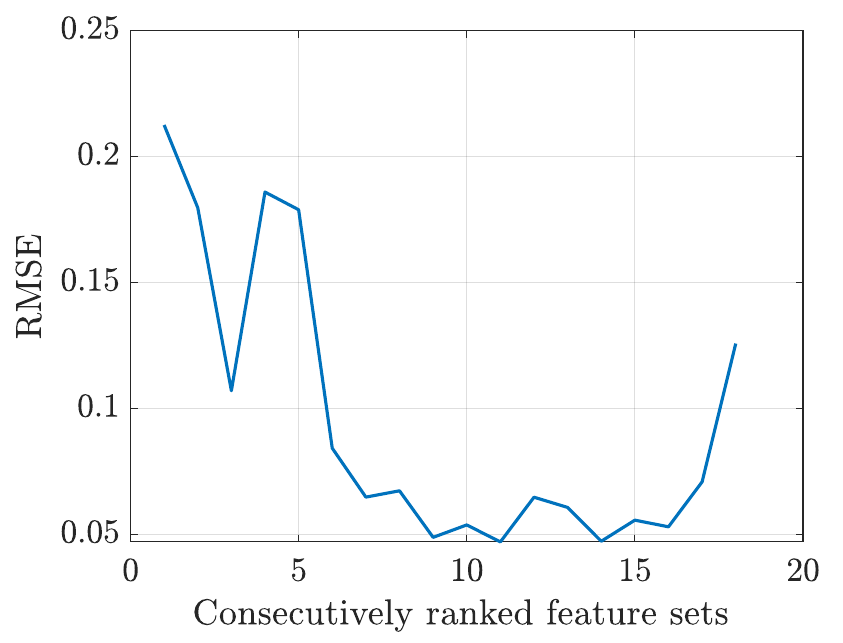}}
  \subfigure[]{\includegraphics[scale=.3]{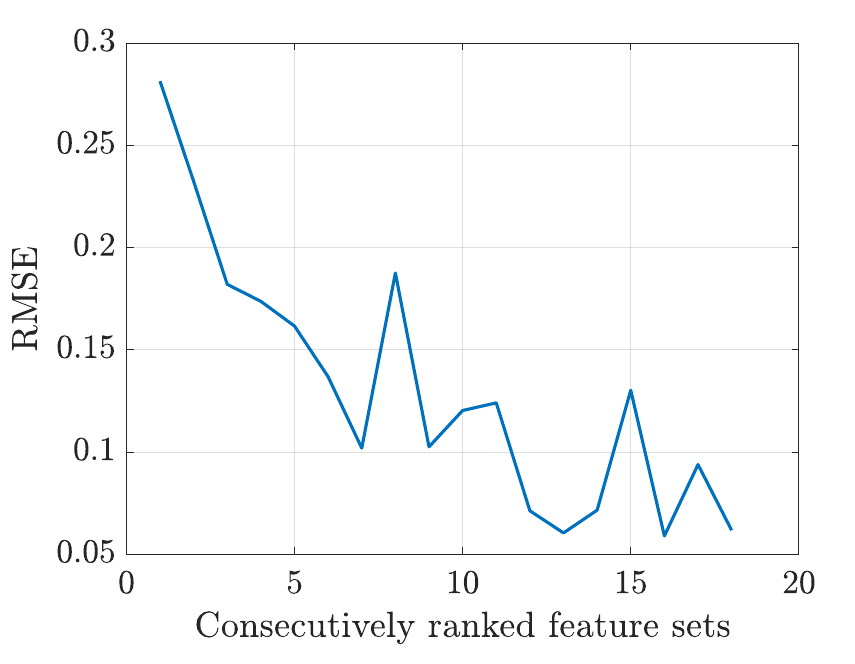}}
  \subfigure[]{\includegraphics[scale=.3]{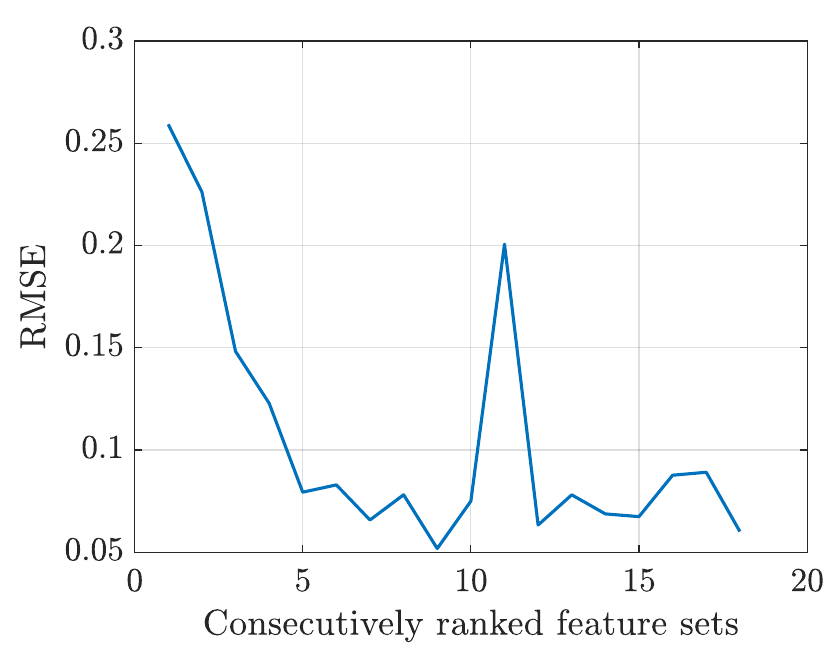}}
  \subfigure[]{\includegraphics[scale=.3]{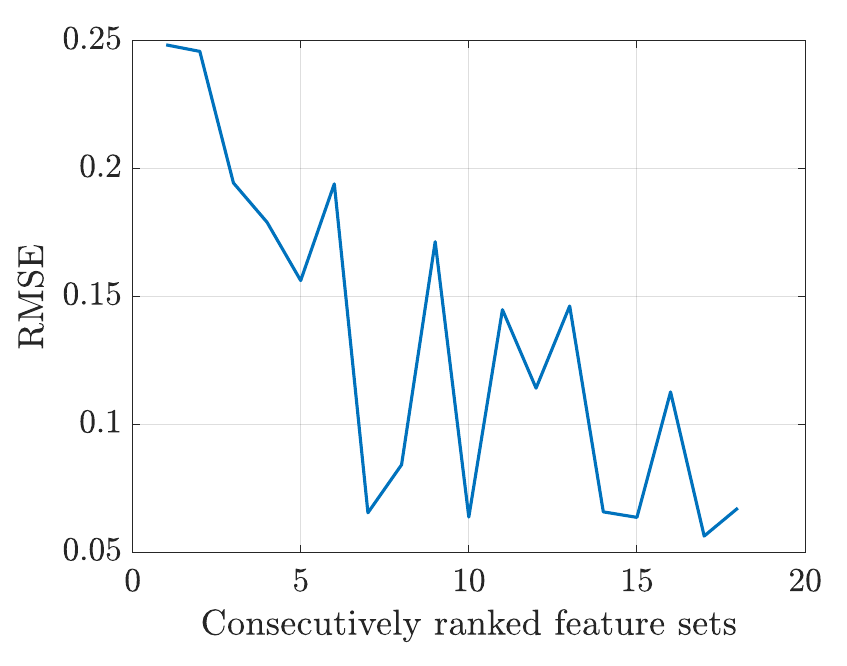}}
  \subfigure[]{\includegraphics[scale=.3]{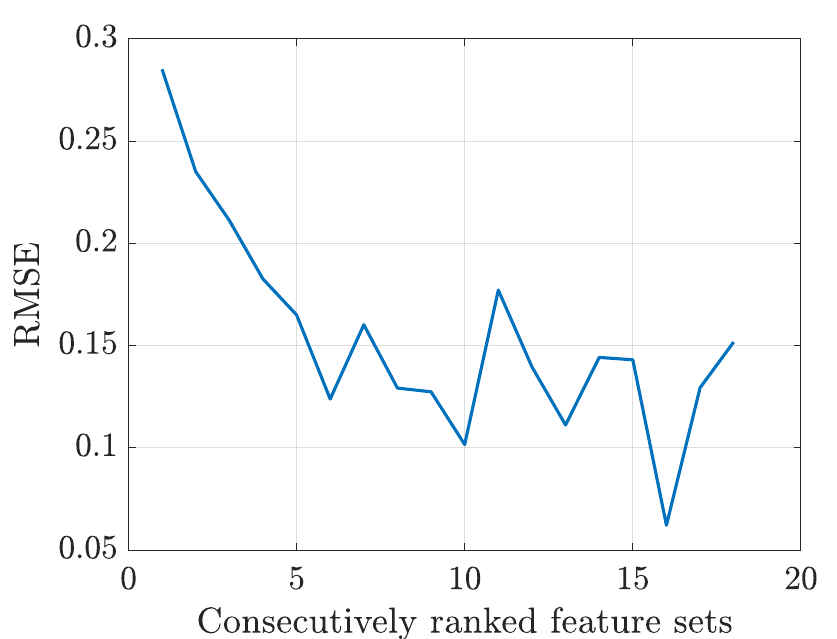}}
\caption{Forward search for predicting damping $d$ using ranked features: (a) $\delta = 1.0$, (b) $\delta = 1.2$, (c) $\delta = 1.4$, (d) $\delta = 1.6$, (e) $\delta = 1.8$, (f) $\delta = 2.0$}
\label{fig:search_d}
\end{figure}

\begin{figure}[!ht]
\centering
  \subfigure[]{\includegraphics[scale=.3]{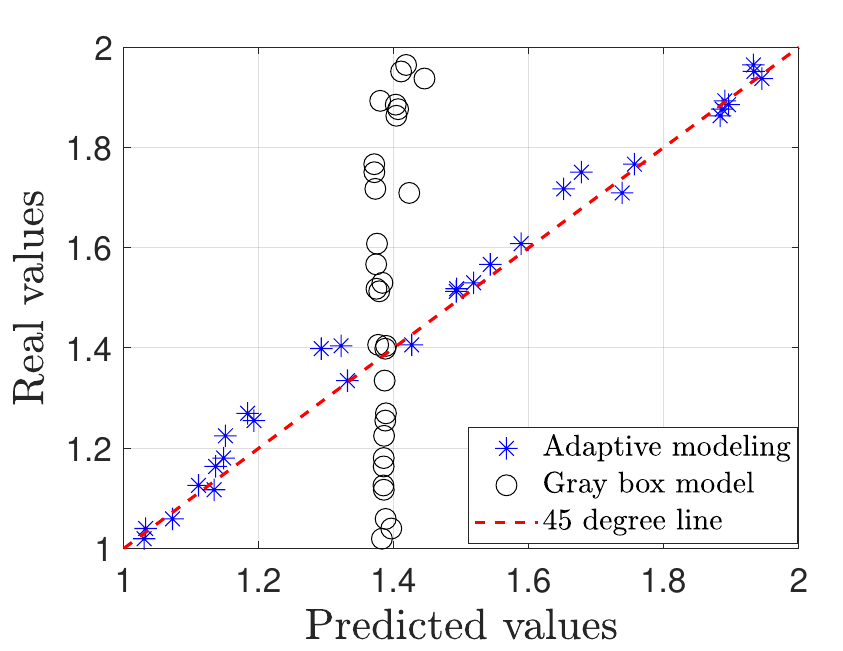}}
  \subfigure[]{\includegraphics[scale=.3]{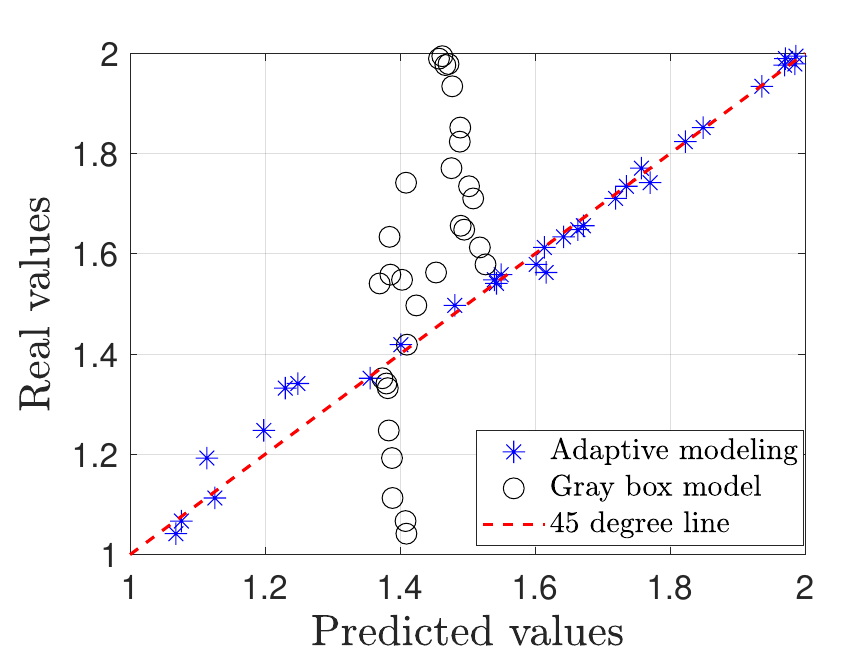}}
  \subfigure[]{\includegraphics[scale=.3]{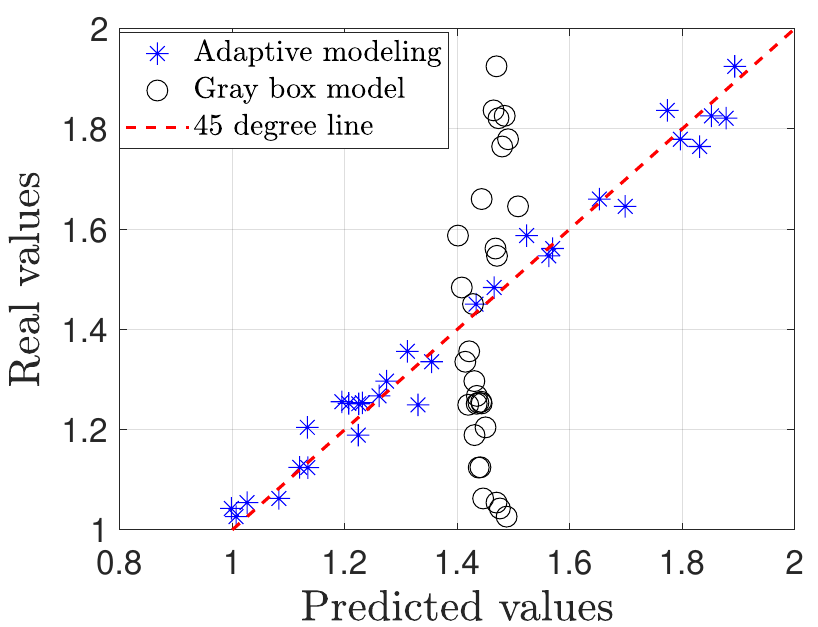}}
  \subfigure[]{\includegraphics[scale=.3]{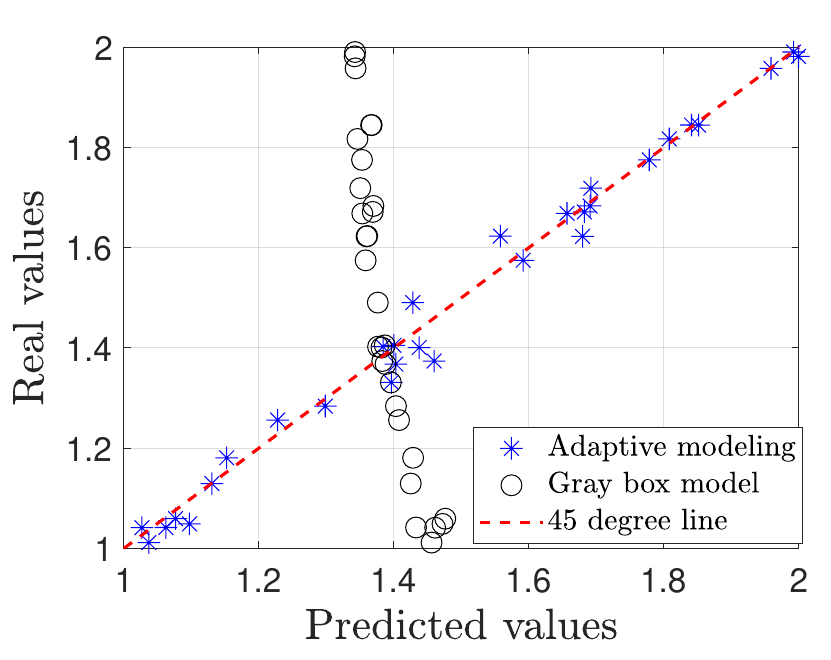}}
  \subfigure[]{\includegraphics[scale=.3]{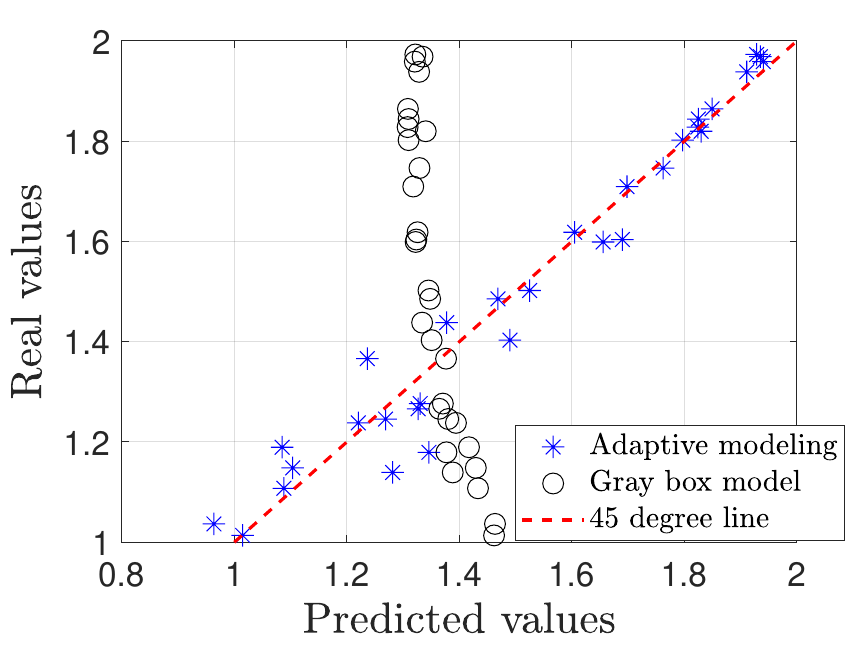}}
  \subfigure[]{\includegraphics[scale=.3]{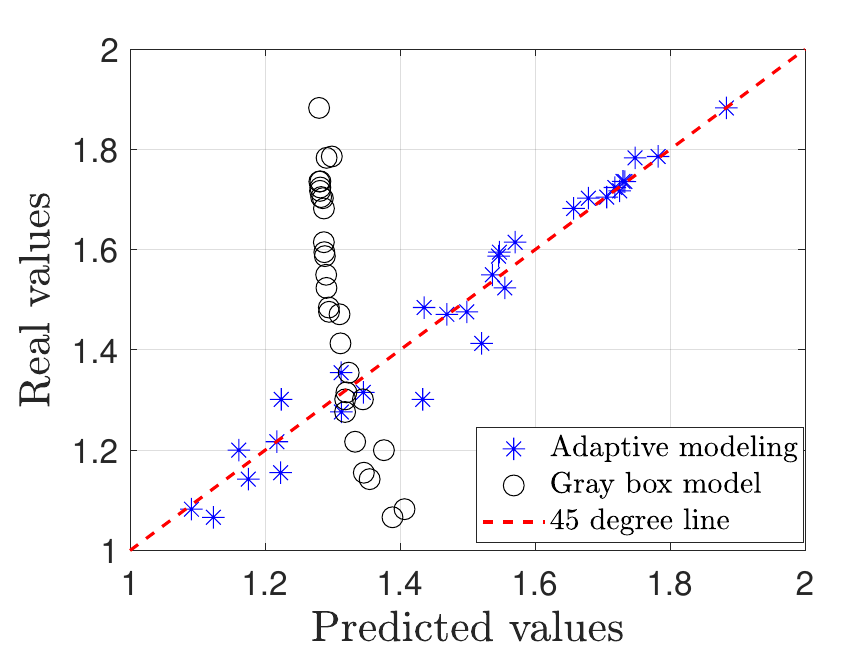}}
  \caption{Comparison of adaptive model and gray box model for predicting damping $d$: (a) $\delta=1.0$, (b) $\delta=1.2$, (c) $\delta=1.4$, (d) $\delta=1.6$, (e) $\delta=1.8$, (f) $\delta=2.0$}
\label{fig:Compare_d}
\end{figure}

Therefore, a more sophisticated feature set is required in this situation and further analysis is also necessary at this point. Fig.~\ref{fig:search_d} provides the results of the forward search which offers insightful information on all features. It can be observed from Fig.~\ref{fig:search_d} that, for scenarios $(a), (b), (d)$, and $(f)$  the participation of the top six consecutive ranked features reduces the RMSE substantially in the forward search, whereas with the inclusion of other features the RMSE of forward search fluctuates. This indicates that the top six features significantly influence the estimation of damping for the above scenarios. These particular features are identified as amplitude-related features with higher MI as shown in Figs.~\ref{fig:rank_d} $(a), (b), (d)$, and $(f)$ respectively. However, for scenarios $(c)$ and $(e)$ in Fig.~\ref{fig:search_d}, the involvement of the top two and top five consecutively ranked features results in the reduction of RMSE during the forward search, while the inclusion of other features leads to RMSE fluctuations in both cases. For scenario $(c)$ the top two features are identified as frequency related as shown in Fig.~\ref{fig:rank_d} $(c)$. For scenario $(e)$ the top five features include two frequency-related features and three amplitude-related features as depicted in Fig.~\ref{fig:rank_d} $(e)$. As a result, it is better to focus on amplitude-related features and the dominant frequency-related features $f_{11}$ and $f_{21}$ for developing a feature set. Finally, all amplitude features $p_{11},p_{12},p_{13},p_{14},p_{21},p_{22},p_{23},p_{24}$ together with the two frequency features $f_{11}$ and $f_{21}$ are chosen as the feature set to train the ANN in this case. Subsequently, repeating the above procedure, the test data is generated from other sources and is fed to the trained neural network to test the predictions of damping coefficient $d$. The gray box modeling is also carried out to estimate the unknown damping $d$ for comparison. Each time series data is generated for varying excitation $f$ and damping coefficient $d$ and with a constant nonlinear stiffness. These data sets are then generated for six scenarios of $\delta$ as in the case of the proposed technique above and the damping coefficient $d$ is estimated. Note that in this case, the regularization technique is used to ensure the convergence of the gray box model. The results from the proposed adaptive technique are compared with the estimation of the gray box model as shown in
Fig. \ref{fig:Compare_d}.

\begin{table}[t] \small
	\centering

	\begin{tabular}{ c | c | c | c }
 \hline \hline
		\textbf{Scenario} & $\delta$ &  \ZL{$\mathrm{RMSE}_a$} & \ZL{$\mathrm{RMSE}_g$} \\
  \hline \hline
   1 & $1.0$ & $0.0419$ & $0.3200$\\
   \hline
   2 & $1.2$ &  $0.0341$ & $0.2851$\\
   \hline
   3 & $1.4$ & $0.0399$ & $0.2665$\\
   \hline
   4 & $1.6$ & $0.0334$ & $0.3539$\\
   \hline
   5 & $1.8$ & $0.0630$  & $0.3808$\\
   \hline
   6 & $2.0$ & $0.0456$  & $0.3167$\\ 	
   \hline
	\end{tabular}
 \caption{Performance comparison between gray box model and adaptive modeling for predicting damping with different scenarios (\ZL{$\mathrm{RMSE}_a$}: the root mean square error of estimations from the hybrid adaptive modeling, \ZL{$\mathrm{RMSE}_g$}: the root mean square error of estimations from the gray box modeling)}
 \label{table:damping}
\end{table}

The performances of both adaptive modeling and the gray box model are evaluated by \ZL{$\mathrm{RMSE}_{a}$} and \ZL{$\mathrm{RMSE}_{g}$} for each scenario demonstrated in Table.~\ref{table:damping}. It is observed that the accuracy of the proposed technique in predicting damping is higher than the gray box model. Note that the proposed adaptive modeling technique predicts the damping coefficient accurately for all scenarios of coupling coefficient. However, the gray box only predicts the values close to the initial guess.

\ZL{The generation of training sets for the adaptive model also required approximately $2500$ seconds and another $85$ seconds for model training. The gray-box model required an average of $250$ seconds for each new estimation. For the $40$ estimations corresponding to the testing set, the trained adaptive model spent $18$ minutes while the gray-box model took more than $30$ hours.}

To summarize this section, the proposed hybrid adaptive modeling method is applied in more realistic situations to track both the coupling and damping coefficients. The higher accuracy of the proposed adaptive model than the gray box model in diverse situations illustrates its robustness, applicability, and generalizability. \ZL{In addition, the time and computational cost of developing the adaptive modeling are highly favorable compared to other machine learning models in similar domains. Moreover,  the adaptive modeling method demonstrates a substantial advantage over the gray box model in prediction with the trained model.}

In comparison, the time and computational costs associated with developing the adaptive model are significantly more favorable than those of comparable machine learning models. Furthermore, regarding prediction speed, the adaptive model demonstrates a substantial advantage over the gray box model.

\section{Conclusion}
\label{se:conclusion}

In this paper, we developed a hybrid model that combines nonlinear dynamics and data-based modeling to adapt to the changes in parameters of a dynamical system. The proposed adaptive model is demonstrated using a set of two coupled Duffing oscillators. The frequency response curves are obtained analytically by the method of multiple scales (MMS) for different parameters. These analytical solutions are used to generate synthetic data considering different scenarios that sufficiently represent the experimental data. An Artificial Neural Network is trained and tested by extracting features from frequency response curves. The features are selected and ranked using mutual information (MI) and forward search to create appropriate feature sets. The estimations from the proposed adaptive technique are compared with the estimations from the gray box model.

\ZL{To demonstrate the generalizability of the proposed method, different scenarios with varying parameters, random loads and added noise which represent the uncertainties in real-world situations are employed. In general, the proposed hybrid adaptive model proved to be superior to the gray box modeling, purely numerical modeling, in accuracy and computational time in estimating desired parameters. For estimating coupling coefficients, the average RMSEs of the adaptive model and the gray box model are $0.0098$ and $0.1164$ separately, the performance superiority is quantified as a $90.85\%$ reduction in RMSE and $95.84\%$ less in computational time. For damping estimation, with an average RMSE reduction from $0.3205$ to $0.0374$, the adaptive model achieves a lower RMSE of nearly $88.33\%$ and a lower computational time of nearly $96.58\%$.} The obtained results demonstrate the effectiveness of the dynamics-based features in capturing the nonlinear dynamics and tracking varying parameters. Moreover, the synthetic data generated by the asymptotic solution compensates for the deficiency of high-quality data in specific circumstances.

This study offers a fresh perspective on extracting dynamics-based features from analytical solutions which are more informative than numerical simulations. These features are better at capturing phenomena of dynamics in nonlinear systems. \ZL{To our knowledge, this is the first time that anyone has explored the integration of deep insights gained from the science of nonlinear dynamics directly into machine learning. The insight gained from this paper would be certainly useful for practical design because all practical designs are nonlinear, and nonlinear phenomena are always observed in practice. It would be especially of practical relevance in systems with harmonic excitation which includes most mechanical and electromechanical systems}

Our future work focuses on extracting new dynamics-based features from the frequency response. Furthermore, expanding the use of the perturbation methods to evaluate superharmonic and subharmonic responses, as well as secondary resonance, may lead to more informative features. 

\section*{Acknowledgments}
This project was supported by research grants from the Office of Naval Research (Grant No. N00014-19-1-2070 and N00014-22-1-2480). We are grateful to ONR and Capt. Lynn Petersen, the program manager for recognizing the importance of this work.


\appendix
\section{}
\label{app1}

Rewriting the \eq{eq:order0} in the form of a matrix, we get the following equations.
\begin{equation}
    \begin{split}
        \begin{bmatrix} 1 & 0 \\ 0 & 1\end{bmatrix}
        \begin{bmatrix} D_0^{2} x_{10}\\D_0^{2} x_{20}\end{bmatrix} +
        \begin{bmatrix}  \omega_{0}^{2} + \delta & -\delta\\
        -\delta & \omega_{0}^{2} + \delta \end{bmatrix}
        \begin{bmatrix} x_{10}\\x_{20}\end{bmatrix} =
        \begin{bmatrix}
        0\\0
        \end{bmatrix}
    \end{split}
    \label{eq:Matrix of order0}
\end{equation}

We assume an exponential solution of \eq{eq:Matrix of order0}.
\begin{equation}
    \begin{split}
        \begin{bmatrix}
        x_{10}\\x_{20}
        \end{bmatrix}
        =
        \begin{bmatrix}
        u_{n1}\\u_{n2}
        \end{bmatrix} e^{st}
    \end{split}
    \label{eq:order0_2}
\end{equation}
Substituting it into \eq{eq:Matrix of order0}, we obtain the equations as follows.
\begin{eqnarray}
    \begin{split}
        (s^2\begin{bmatrix} 1 & 0 \\ 0 & 1 \end{bmatrix}
         +
        \begin{bmatrix}  \omega_{0}^{2} + \delta & -\delta\\
        -\delta & \omega_{0}^{2} + \delta \end{bmatrix}
        ) \begin{bmatrix} u_{n1}\\u_{n2} \end{bmatrix} e^{st} =
        \begin{bmatrix} 0\\0\end{bmatrix}
    \end{split}
    \label{eq:order0_3}
\end{eqnarray}

Since $e^{st}\neq0$, we get the eigenvalue problem as follows.
\begin{equation}
    \begin{split}
        (s^2\begin{bmatrix} 1 & 0 \\ 0 & 1 \end{bmatrix}+
        \begin{bmatrix} \omega_{0}^{2} + \delta & -\delta\\
        -\delta & \omega_{0}^{2} + \delta \end{bmatrix})
        \begin{bmatrix} u_{n1}\\u_{n2}\end{bmatrix}=
        \begin{bmatrix} 0\\0\end{bmatrix}
    \end{split}
    \label{eq:order0_3_2}
\end{equation}
By introducing $\omega_{n} = i s$, $\omega_{n}^2 = -s^{2}$, we obtain the following equation.
\begin{equation}
    \begin{split}
    \begin{vmatrix}
    \omega_{0}^{2} + \delta - \omega_{n}^2 & -\delta\\
    -\delta & \omega_{0}^{2} + \delta - \omega_{n}^2
    \end{vmatrix} = 0
    \end{split}
    \label{eq:order0_4}
\end{equation}
The solution of \eq{eq:order0_4} yields two roots.
\begin{equation}
\omega_1 = \omega_0, u_1 = \begin{bmatrix} 1\\1 \end{bmatrix}; \omega_2 = \sqrt{\omega_{0}^2+2\delta},u_2 = \begin{bmatrix} 1\\-1
\end{bmatrix}
\label{eq:order0_5}
\end{equation}
So the solution of \eq{eq:Matrix of order0} has the following form.
\begin{equation}
\begin{split}
x_{10} = a_1\cos{(\omega_1 t + \beta_1)} + a_2\cos{(\omega_2 t + \beta_2)}\\
x_{20} = a_1\cos{(\omega_1 t + \beta_1)} - a_2\cos{(\omega_2 t + \beta_2)}
\end{split}
\label{eq:solution1_order0}
\end{equation}
\label{se:appendix}

The simplified solution for homogeneous second-order differential equations in \eq{eq:order0} are as follows.
\begin{equation}
\begin{split}
x_{10} = A_1 e^{i \omega_1 T_0} + A_2 e^{i \omega_2 T_0 } + cc\\
x_{20} = A_1 e^{i\omega_1 T_0} - A_2 e^{i \omega_2 T_0 } + cc
\end{split}
\label{eq:solution2_order0}
\end{equation}

where, $cc$ denotes conjugate complex, $A_1 = \frac{1}{2}a_{1}e^{i\beta_{1}}$, $A_2 = \frac{1}{2}a_{2}e^{i\beta_{2}}$, $\bar{A_1}$ and $\bar{A_2}$ are the complex conjugates of $A_1$ and $A_2$.

Introducing the detuning parameters $\sigma_1$, $\sigma_2$ as follows.

\begin{equation}
\Omega = \omega_1 + \epsilon \sigma_1, \omega_2 = \omega_1 + \epsilon \sigma_2
\label{eq:detuning}
\end{equation}

Substituting \eq{eq:solution2_order0} and \eq{eq:detuning} into \eq{eq:order1} yields the following form.

\begin{eqnarray}
\begin{split}
&D_0^{2} x_{11} + \omega_0^{2}x_{11} + \delta x_{11} - \delta x_{21} =  \\& (-2D_1 A_1 i \omega_1 - d A_1 i \omega_1 -3\beta {A_1}^2 \Bar{A_1}  - 6 \beta A_1 A_2 \Bar{A_2} + 0.5 f e^{i \sigma_1 T_1})e^{i \omega_1 T_0} \\& + (-2D_1 A_2 i \omega_2 - d A_2 i \omega_2  -3\beta {A_2}^2 \Bar{A_2} - 6 \beta A_1 A_2 \Bar{A_1}) e^{i \omega_2 T_0}  \\ & -3\beta {A_1}^2 A_2 e^{(2 \omega_1 + \omega_2)i T_0} - 3\beta {A_1}^2 \bar{A_2} e^{(2 \omega_1 - \omega_2)i T_0} \\ & -3\beta A_1 {A_2}^2 e^{(\omega_1 + 2 \omega_2)i T_0} - 3\beta \bar{A_1} {A_2}^2 e^{(2 \omega_2 - \omega_1)i T_0} \\& D_0^{2} x_{21} + \omega_0^{2}x_{21} + \delta x_{21} - \delta x_{11}= \\& (-2D_1 A_1 i \omega_1 - d A_1 i \omega_1 -3\beta {A_1}^2 \Bar{A_1} - 6 \beta A_1 A_2 \Bar{A_2})e^{i \omega_1 T_0} \\&+ (2D_1 A_2 i \omega_2 + d A_2 i \omega_2 +3\beta {A_2}^2 \Bar{A_2} + 6 \beta A_1 A_2 \Bar{A_1}) e^{i \omega_2 T_0} \\ &+3\beta {A_1}^2 A_2 e^{(2 \omega_1 + \omega_2)i T_0} + 3\beta {A_1}^2 \bar{A_2} e^{(2 \omega_1 - \omega_2)i T_0} \\& -3\beta A_1 {A_2}^2 e^{(\omega_1 + 2 \omega_2)i T_0} - 3\beta \bar{A_1} {A_2}^2 e^{(2 \omega_2 - \omega_1)i T_0}
\label{eq:Primary}
\end{split}
\end{eqnarray}

We assume the solution of the homogeneous form for \eq{eq:Primary} is as follows.

\begin{eqnarray}
\begin{split}
x_{11} &= B_{11} e^{i \omega_1 T_0} + B_{12} e^{i \omega_2 T_0 } + cc,\\
x_{21} &= B_{21} e^{i \omega_1 T_0} + B_{22} e^{i \omega_2 T_0 } + cc.
\label{eq:solution1_order1}
\end{split}
\end{eqnarray}

Substituting the \eq{eq:solution1_order1} into \eq{eq:Primary}, and collecting the secular terms we get the following equations.
\begin{eqnarray}
\begin{split}
R_{11} e^{i \Omega_1 T_0} + R_{12} e^{i \omega_2 T_0}  = &(-{\omega_1}^2 + \omega_{0}^2 + \delta)B_{11} e^{i \omega_1 T_0} - \delta B_{21}e^{i \omega_1 T_0} \\&+ (-{\omega_2}^2 + \omega_{0}^2 + \delta) B_{12} e^{i \omega_2 T_0}  - \delta B_{22} e^{i \omega_2 T_0} \\ R_{21} e^{i \omega_1 T_0} + R_{22} e^{i \omega_2 T_0} =& (-{\omega_1}^2 + \omega_{0}^2 + \delta)B_{21} e^{i \omega_1 T_0} - \delta B_{11}e^{i \omega_1 T_0} \\&+ (-{\omega_2}^2 + \omega_{0}^2 + \delta) B_{22} e^{i \omega_2 T_0} - \delta B_{12} e^{i \omega_2 T_0},
\label{eq:secular}
\end{split}
\end{eqnarray}

where,
\begin{eqnarray}
\begin{split}
R_{11} =& -2 D_1 A_{1} i \omega_1 - 2 D_1 A_{2}i \omega_2 e^{i \sigma_2 T_1} - d A_1 i \omega_1 r- d A_2 i \omega_2 e^{i \sigma_2 T_1}  \\& - 3 \beta {A_1}^2 \Bar{A_1} - 3 \beta {A_2}^2 \Bar{A_2} e^{i \sigma_2 T_1} - 6 \beta A_1 A_2 \Bar{A_2} \\&- 6 \beta A_1 \Bar{A_1} A_2 e^{i \sigma_2 T_1} + \frac{1}{2} f e^{i \sigma_1 T_1} \\
R_{12} =& -2 D_1 A_{1} i \omega_1 e^{-i \sigma_2 T_1} - 2 D_1 A_{2}i \omega_2 - d A_1 i \omega_1 e^{-i \sigma_2 T_1} - d A_2 i \omega_2  \\&- 3 \beta {A_1}^2 \Bar{A_1} e^{-i \sigma_2 T_1} - 3 \beta {A_2}^2 \Bar{A_2} - 6 \beta A_1 A_2 \Bar{A_2} e^{-i \sigma_2 T_1} \\&- 6 \beta A_1 \Bar{A_1} A_2 + \frac{1}{2} f e^{i (\sigma_1 -\sigma_2) T_1} \\R_{21} =& -2 D_1 A_{1} i \omega_1 + 2 D_1 A_{2}i \omega_2 e^{i \sigma_2 T_1} - d A_1 i \omega_1 \\&+ d A_2 i \omega_2 e^{i \sigma_2 T_1} - 3 \beta {A_1}^2 \Bar{A_1} + 3 \beta {A_2}^2 \Bar{A_2} e^{i \sigma_2 T_1} \\&- 6 \beta A_1 A_2 \Bar{A_2} + 6 \beta A_1 \Bar{A_1} A_2 e^{i \sigma_2 T_1} \\R_{22} =& -2 D_1 A_{1} i \omega_1 e^{-i \sigma_2 T_1} + 2 D_1 A_{2}i \omega_2 - d A_1 i \omega_1 e^{-i \sigma_2 T_1} \\&+ d A_2 i \omega_2 - 3 \beta {A_1}^2 \Bar{A_1} e^{-i \sigma_2 T_1} + 3 \beta {A_2}^2 \Bar{A_2} \\&- 6 \beta A_1 A_2 \Bar{A_2} e^{-i \sigma_2 T_1} + 6 \beta A_1 \Bar{A_1} A_2.
\label{eq:R1_R2}
\end{split}
\end{eqnarray}
\eq{eq:secular} can also be written in a matrix form as follows.

\begin{equation}
    \begin{split}
        \begin{bmatrix} -{\omega_1}^2+{\omega_0}^2 - \delta & -\delta \\ -\delta & -{\omega_1}^2+{\omega_0}^2 + \delta
        \end{bmatrix} \begin{bmatrix} B_{11}\\B_{21} \end{bmatrix}= \begin{bmatrix} R_{11}\\R_{21} \end{bmatrix}\\
        \begin{bmatrix} -{\omega_2}^2+{\omega_0}^2 - \delta & -\delta \\ -\delta & -{\omega_2}^2+{\omega_0}^2 + \delta
        \end{bmatrix} \begin{bmatrix} B_{12}\\B_{22} \end{bmatrix}= \begin{bmatrix} R_{12}\\R_{22} \end{bmatrix}\\
    \end{split}
    \label{eq:Matrix}
\end{equation}
Substituting $\omega_1 = \omega_0$, $\omega_2=\sqrt{\omega_{0}^2 + 2\delta}$ into \eq{eq:Matrix}, yields the solvability conditions in terms of $R_{1n}$ and $R_{2n}$.
\begin{equation}
\begin{split}
R_{11}+R_{21} = 0\\
-R_{12}+R_{22} = 0
\end{split}
\label{eq:Solvability}
\end{equation}
Substituting \eq{eq:R1_R2} into \eq{eq:Solvability}, the solvability conditions reduce to $D_1 A_1$ and $D_1A_2$ as follows.
\begin{eqnarray}
\begin{split}
        -4D_1 A_1i \omega_1 -2d A_1i \omega_1 - 6\beta A_{1}^2 \bar{A_1} - 12\beta A_1 A_2 \bar{A_2} + 0.5 f e^{i \sigma_1 T_1} =&0 \\
        4D_1 A_2i \omega_2 +2d A_2i \omega_2 + 6\beta A_{2}^2 \bar{A_2} + 12\beta A_1 A_2 \bar{A_1} - 0.5 f e^{i (\sigma_1 - \sigma_2) T_1} =&0
\end{split}
\label{eq:A1_A2}
\end{eqnarray}

Substituting $A_1 = \frac{1}{2}a_{1}e^{i\beta_{1}}$, $A_2 = \frac{1}{2}a_{2}e^{i\beta_{2}}$ into \eq{eq:A1_A2} yields the following equations.
\begin{eqnarray}
\begin{split}
    2i\omega_{1} a_1' + i\omega_1 d a_1 - \frac{1}{2}if \sin{(\sigma_1 T_1 - \beta_1)} - \\2 \omega_1 a_1 \beta_1'  + \frac{3}{4} \beta a_{1}^3 + \frac{3}{2} \beta a_1 a_{2}^2 - \frac{1}{2} f \cos{(\sigma_1 T_1 - \beta_1)} = &0 \\ 2i\omega_{2} a_2' + i\omega_2 d a_2 - \frac{1}{2}if \sin{(\sigma_1 T_1 - \sigma_2 T_1 - \beta_2)} \\- 2 \omega_2 a_2 \beta_2'  + \frac{3}{4} \beta a_{2}^3 + \frac{3}{2} \beta a_2 a_{1}^2 - \frac{1}{2} f \cos{(\sigma_1 T_1 -\sigma_2 T_1- \beta_2)} =& 0
    \label{eq:der_modulation}
    \end{split}
\end{eqnarray}
Separating real and imaginary terms, we obtain the modulation equations.

\begin{eqnarray}
\begin{split}
a_1' &= -\frac{1}{2} d a_1 + \frac{1}{4 \omega_1} f \sin({\sigma_1 T_1 - \beta_1}) \\
a_1 \beta_1' &= \frac{3}{8 \omega_1} \beta a_1^3 + \frac{3}{4 \omega_1} \beta a_1 a_2^2 - \frac{1}{4 \omega_1} f \cos({\sigma_1 T_1 -\beta_1}) \\
a_2' &= -\frac{1}{2} d a_2 + \frac{1}{4 \omega_2} f \sin({\sigma_1 T_1 - \sigma_2 T_1 - \beta_2}) \\
a_2 \beta_2' &= \frac{3}{8 \omega_2} \beta a_2^3 + \frac{3}{4 \omega_2} \beta a_2 a_1^2 - \frac{1}{4 \omega_2} f \cos({\sigma_1 T_1 - \sigma_2 T_1 - \beta_2})
\label{eq:modulation1}
\end{split}
\end{eqnarray}

Subsequently, we obtain the following set of autonomous forms of modulation equations.
\begin{eqnarray}
\begin{split}
a_1' &= -\frac{1}{2} d a_1 + \frac{1}{4 \omega_1} f \sin(\gamma_1)  \\
\gamma_1' &= \sigma_1 - \frac{3}{8 \omega_1} \beta a_1^2 - \frac{3}{4 \omega_1} \beta a_2^2 + \frac{1}{4 \omega_1 a_1} f \cos(\gamma_1) \\
a_2' &= -\frac{1}{2} d a_2 + \frac{1}{4 \omega_2} f \sin(\gamma_2) \\
\gamma_2' &= \sigma_1 - \sigma_2 - \frac{3}{8 \omega_2} \beta a_2^2 - \frac{3}{4 \omega_2} \beta a_1^2 + \frac{1}{4 \omega_2 a_2} f \cos(\gamma_2),
\label{eq:modulation2}
\end{split}
\end{eqnarray}
where,
\begin{equation}
\gamma_1 = \sigma_1 T_1 -\beta_1, \gamma_2 = \sigma_1 T_1 -\sigma_2 T_1 -\beta_2
\label{eq:gamma}
\end{equation}

By solving ~\eq{eq:modulation2} and substituting the obtained $a_{1}$, $\gamma_1$, $a_2$, $\gamma_2$,~\eq{eq:gamma} and ~\eq{eq:detuning} into~\eq{eq:solution2_order0}, we obtain the following equation.
\begin{equation}
\begin{split}
x_{10} &= a_1\cos{(\omega_1 t + \sigma_1 T_1 -\gamma_1)} + a_2\cos{(\omega_1 t + \sigma_1 T_1 -\gamma_2)}\\
x_{20} &= a_1\cos{(\omega_1 t + \sigma_1 T_1 -\gamma_1)} - a_2\cos{(\omega_1 t + \sigma_1 T_1 -\gamma_2)}
\end{split}
\label{eq:solution1_order0_newform}
\end{equation}

By using trigonometric identities, \eq{eq:solution1_order0_newform} is rewritten in the following form.

\begin{eqnarray}
\begin{split}
x_{10} =&  (a_1\cos{(\omega_1 t + \sigma_1 T_1)}\cos{\gamma_1} + \sin{(\omega_1 t + \sigma_1 T_1)}\sin{\gamma_1}) \\&+ (a_2\cos{(\omega_1 t + \sigma_1 T_1)}\cos{\gamma_2} + \sin{(\omega_1 t + \sigma_1 T_1)}\sin{\gamma_2})  \\
x_{20} =&  (a_1\cos{(\omega_1 t + \sigma_1 T_1)}\cos{\gamma_1} + \sin{(\omega_1 t + \sigma_1 T_1)}\sin{\gamma_1}) \\&- (a_2\cos{(\omega_1 t + \sigma_1 T_1)}\cos{\gamma_2} + \sin{(\omega_1 t + \sigma_1 T_1)}\sin{\gamma_2})
\label{eq:angle sum}
\end{split}
\end{eqnarray}





 \bibliographystyle{elsarticle-num}
 \bibliography{main}



\end{document}